\RequirePackage[svgnames]{xcolor}

\documentclass[11pt,letterpaper,logo]{mystyle}

\usepackage[all]{hypcap}
\usepackage[svgnames]{xcolor}
\usepackage[comma,authoryear,compress]{natbib}
\usepackage{hyperref}[citecolor=lightblue]

\hypersetup{
    colorlinks = true,
    citecolor = {MSAccent},
}

\usepackage{algorithm}
\usepackage{algorithmicx}
\usepackage{algpseudocode}
\usepackage{microtype}
\usepackage{graphicx}
\expandafter\def\csname ver@subfig.sty\endcsname{}
\usepackage{booktabs} %
\usepackage{float}
\usepackage{bigstrut}

\usepackage{amsmath}
\usepackage{amssymb}
\usepackage{mathtools}
\usepackage{amsthm}
\usepackage{mathrsfs}
\usepackage{nicefrac}
\usepackage{dsfont}
\usepackage{enumitem}
\usepackage{subcaption}
\usepackage{graphicx,subfig}
\usepackage{cleveref}

\usepackage{float}

\usepackage[utf8]{inputenc} %
\usepackage[T1]{fontenc}    %
\usepackage{hyperref}       %
\usepackage{url}            %
\usepackage{booktabs}       %
\usepackage{amsfonts}       %
\usepackage{nicefrac}       %
\usepackage{microtype}      %
\usepackage{graphicx}
\usepackage{subcaption} 
\usepackage{amssymb}
\usepackage{fdsymbol}
\usepackage{wrapfig}
\usepackage{lipsum}
\usepackage{enumitem}
\usepackage{stackengine}
\usepackage[font=small,labelfont=bf]{caption}
\usepackage{color}
\usepackage{adjustbox}

\usepackage{rotating}
\usepackage{makecell}

\definecolor{blanchedalmond}{rgb}{1.0, 0.92, 0.8}
\definecolor{carmine}{rgb}{0.59, 0.0, 0.09}
\definecolor{lightblue}{rgb}{0.22,0.45,0.70}%

\renewcommand{\mathbf}{\boldsymbol}

\makeatletter
\def\Ddots{\mathinner{\mkern1mu\raise\p@
\vbox{\kern7\p@\hbox{.}}\mkern2mu
\raise4\p@\hbox{.}\mkern2mu\raise7\p@\hbox{.}\mkern1mu}}
\makeatother

\definecolor{amaranth}{rgb}{0.9, 0.17, 0.31}
\definecolor{antiquebrass}{rgb}{0.8, 0.58, 0.46}
\definecolor{antiquefuchsia}{rgb}{0.57, 0.36, 0.51}
\definecolor{chromeyellow}{rgb}{0.31, 0.47, 0.26}

\newcommand{\github}{\raisebox{-1.5pt}{\includegraphics[height=1.05em]{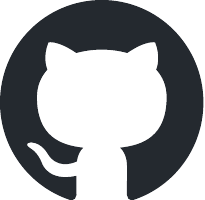}}}
\newcommand{\mslogo}{\raisebox{-1.5pt}{\includegraphics[height=1.05em]{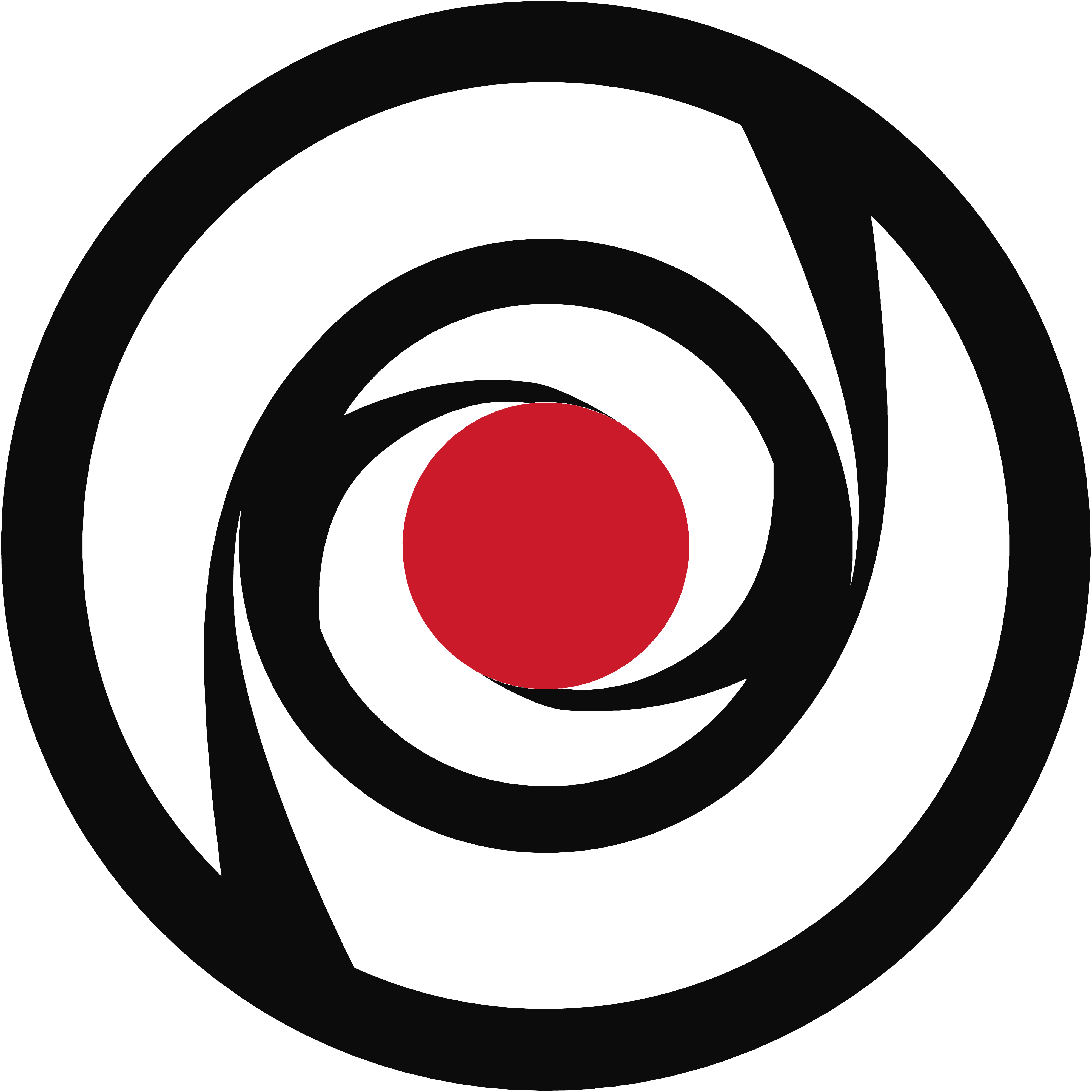}}}
\newcommand{\hflogo}{\raisebox{-1.5pt}{\includegraphics[height=1.05em]{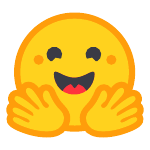}}}

\usepackage{pifont}
\usepackage{xspace}
\usepackage{listings}
\newcommand{\yes}{\ding{51}}
\newcommand{\no}{\ding{55}}
\newcommand{\partly}{$\triangle$}
\newcommand{\bench}{CameraLogBench\xspace}

\newtcolorbox{AIbox}[2][]{aibox,title=#2,#1}
\definecolor{lightblue}{rgb}{0.22,0.45,0.70}%
\definecolor{Gray}{gray}{0.95}
\definecolor{Cornsilk}{rgb}{1.0, 0.97, 0.86}

\usepackage{amsmath}

\usepackage[all]{hypcap}

\title{MediaStormBench-CameraLog:\\ Can Multimodal Models Identify a Camera's Log Curve?}
\runningtitle{MediaStormBench-CameraLog: Can Multimodal Models Identify a Camera's Log Curve?}

\author[1*\dag]{Ming Ma}
\author[6*]{Wenhao Hu}
\author[2]{Qichao Wang}
\author[3]{Yuhao Wang}
\author[4]{Junpeng Zhao}
\author[5]{Qiyi Wang}
\author[6]{Fangrui Zhang}
\author[6]{Zhiqian Lai}
\author[6]{Shanxin Guan}
\author[6]{Haoyi Pei}
\author[6]{Hongan Chen}
\author[6]{Rui Gong}
\affil[1]{UCAS}
\affil[2]{NTU}
\affil[3]{SJTU}
\affil[4]{FDU}
\affil[5]{NJU}
\affil[6]{MediaStorm}

\authornote{$^*$Equal contribution\quad $^\dag$Project Lead}

\correspondingauthor{Rui Gong (\href{mailto:gary@ysjf.com}{gary@ysjf.com}). Email: Ming Ma (\href{mailto:mam22@ion.ac.cn}{mam22@ion.ac.cn}), Wenhao Hu (\href{mailto:winstonhu@ysjf.com}{winstonhu@ysjf.com})}

\begin{document}

\begin{abstract}
MediaStormBench is a multimodal benchmark built for real-world film and television production workflows, with each subset distilled from practical needs in MediaStorm's day-to-day video production.
This paper releases its first subset, MediaStormBench-CameraLog (CameraLogBench), focusing on a common foundational task in footage ingest and initial color grading: determining which Log curve was used to shoot a clip without metadata.
Cameras record Log footage by compressing a wide range of scene light into pixel code values with a vendor-defined logarithmic curve, and color can be restored correctly in post-production only once that curve is known.
The curve, however, is recorded only in file metadata, which is easily lost in transcoding and in hand-offs between teams.
A colorist is then left to try candidate curves one by one against a waveform scope, and a wrong choice shifts the tones and colors of the entire clip.
Recovering the encoding from the image alone is challenging, because the vendors' curves cluster tightly in code value and are coupled with the unknown exposure.
CameraLogBench is built around professionally shot footage without metadata and tests whether multimodal models can identify a camera's Log curve from the image alone, together with a probe that removes the correct option to expose guessing from priors.
Because every candidate curve is strictly defined by a published vendor formula, we derive four controls from a single implementation of these formulas: graded negatives, synthetic frames of identical scenes, in-context specification digests, and a training-free formula baseline, which together decouple semantic priors from photometric measurement of the pixels.
Across 11 models, accuracy on real footage barely exceeds always guessing the most frequent curve; when the correct option is removed, models almost never abstain, revealing a heavy reliance on scene and brand priors.
Supplying the vendor specification sharply improves recognition on calibrated charts but hardly helps on natural images that lack a photometric reference, exposing a pronounced gap between reading a specification and measuring pixels.

\vspace{4mm}

\mslogo{} \textbf{Project page}: \href{https://tech.ysjf.com/mediastormbench/}{\nolinkurl{https://tech.ysjf.com/mediastormbench}}

\github{} \textbf{Code}: \url{https://github.com/MediaStorm-Tech/MediaStormBench}

\hflogo{} \textbf{Dataset}: \url{https://huggingface.co/datasets/mediatech-tech/MediaStormBench}

\end{abstract}

\maketitle
\vspace{3mm}
\section{Introduction}
\label{sec:intro}

\begin{figure}[t]
  \centering
  \includegraphics[width=\textwidth]{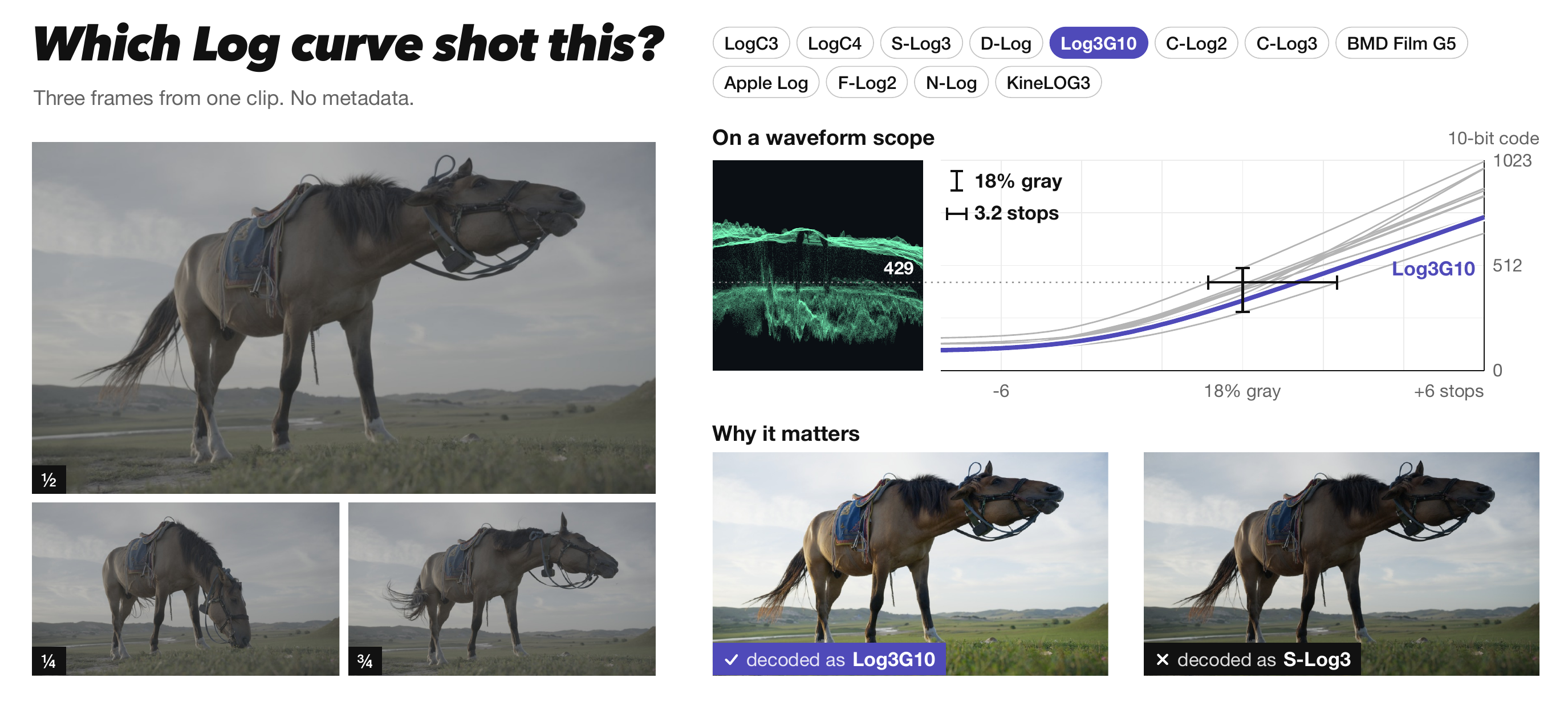}
  \caption{A real benchmark item.
\textbf{Left:} three frames of a Log clip without metadata.
\textbf{Top right:} the 12 official curves; the ground truth is RED Log3G10.
\textbf{Middle right:} the luma waveform of the middle frame, with the frame median 429 dashed, and the 12 curves on the same 10-bit code-value axis. Under these curves an 18\% gray card falls anywhere between code values 285 and 500, and the median 429 decodes to scene light up to 3.2 stops apart.
\textbf{Bottom right:} the frame decoded correctly as Log3G10 and wrongly as S-Log3.}
  \label{fig:example}
\end{figure}

Multimodal models are entering video production workflows.
A team that produces content continuously needs models to take on specific tasks at every stage, from pre-production planning, on-set shooting and footage ingest to editing, color grading and delivery; yet whether a model can actually be integrated into the production pipeline depends on its reliability on that task.
Most existing multimodal benchmarks are designed around general visual understanding and academic reasoning and rarely draw on real production work, so their scores cannot answer this question directly.

We therefore build MediaStormBench.
\href{https://www.ysjf.com/index}{MediaStorm} is a production company that has long published original content on video platforms, and we turn individual steps of its day-to-day production workflow into standardized evaluation subsets.
This paper releases its first subset, \bench, which stems from a recurring, thorny problem in media management and color grading: determining which Log curve was used to shoot a clip without metadata.

Identifying a camera's Log encoding is a prerequisite for color restoration and multi-camera matching in professional post-production.
To preserve a wide scene dynamic range, cinema cameras, drones, mirrorless cameras and phones compress scene light with a vendor-published logarithmic transfer curve paired with a wide gamut, such as Sony S-Log3~\citep{sony2014technical} and ARRI LogC3~\citep{arri2017alexa}.
Unconverted Log footage looks gray and washed out (\Cref{fig:example}, left), and post-production starts by applying the inverse of the curve.
When the wrong curve is applied, the software raises no error, yet the tones and colors of the entire clip are systematically distorted (\Cref{fig:example}, bottom right).
The real frame in \Cref{fig:example} has a median 10-bit code value of 429: decoded as Apple Log it lies 0.9 stops below a standard gray card, whereas decoded as ARRI LogC4 it lies 2.3 stops above---a 3.2-stop (about $9\times$) discrepancy in scene light.
Real productions also routinely mix brands: each of the 12 productions we collected mixes 2 to 4 curves, and a single misidentified camera leaves its shots impossible to align with the others.

The curve in use, however, is recorded only in file metadata, which is easily lost during proxy transcoding, rewrapping and hand-offs between teams.
Editing and grading software relies on this metadata to match the color space: the Premiere Pro documentation notes that transcoded files often have to be tagged by hand~\citep{adobecolor} and that footage whose metadata cannot be read is treated as Rec.709~\citep{adobeauto}.
Faced with gray footage whose metadata is gone, a colorist watches the waveform scope, looks for near-neutral regions and tries the candidate curves one by one, which is slow and relies on experience.
Labeling a long-accumulated archive of untagged footage calls for a method that identifies the Log curve from the image alone.

This paper asks whether multimodal models can recognize a camera's Log curve from frames without metadata.
The task poses two difficulties.
First, the candidate curves cluster tightly in code-value space: of the 12 official curves we consider, 9 map black to the adjacent code values 94 and 95, and 7 map an 18\% gray card into the narrow range from 392 to 420.
Second, the exposure of a shot is unknown: global brightness alone cannot distinguish switching the curve from shifting exposure by a stop.
The answer therefore depends not on which objects the frame shows but on the fine distribution of the pixel code values.
Existing multimodal benchmarks focus on high-level semantic understanding and subject reasoning~\citep{goyal2017vqav2,liu2024mmbench,yue2024mmmu}, on the estimation of capture parameters~\citep{fang2025camerabench} or on the perception of apparent color~\citep{liang2025colorbench}; none tests whether a model can discriminate a camera's opto-electronic encoding from low-level pixel statistics.

The core of \bench is 330 professionally shot clips without metadata that cover all 12 official curves, recorded with 20 camera bodies in total across 12 productions and one dedicated shoot.
Three frames of each clip are given to the model, and we jointly evaluate curve identification, camera-brand identification and a probe with the correct option removed, to tell whether a model is verifying pixels or guessing from priors.
Because each candidate curve is strictly defined by a published vendor formula, we further derive four controls from a single implementation of these formulas: ``not Log'' negatives produced by grading, synthetic frames in which the scene is fixed and only the curve changes, specification digests that the model can consult in context, and an analytic formula baseline that needs no training.

An evaluation of 11 mainstream multimodal models yields three findings.
First, identification on real footage is far from solved: without reference material the best accuracy is only 0.20, while answering D-Log, the most frequent curve, for every item already scores 0.16, and 7 of the 11 models fall below this baseline.
Second, models rely on scene and brand shortcuts: once the correct option is removed they almost never abstain and instead switch to another curve, most often a similar encoding of the same brand.
Third, supplying the vendor specification lifts strong models to 0.82 on calibrated charts yet hardly improves natural images that lack a photometric reference, which exposes a gap between reading a specification and measuring pixels.

Our contributions are as follows:
\begin{enumerate}[leftmargin=*,nosep]
  \item \textbf{The first subset of MediaStormBench and a new task}: formalizing the practical post-production need to determine the Log curve of footage as camera Log curve identification without metadata, with 330 professionally shot clips covering 12 official curves and 20 camera bodies.
  \item \textbf{Four controls derived from vendor formulas}, together with two baselines, which separate content priors from photometric measurement.
  \item \textbf{A diagnosis of 11 multimodal models}, which reveals their bottleneck on real footage, their scene and brand shortcuts, and the gap between calibrated charts and natural images.
\end{enumerate}

\section{Related Work}
\label{sec:related}

\noindent\textbf{Encoding identification and camera provenance.}
Editing and grading software identifies color spaces from file metadata~\citep{adobecolor,adobeauto}.
The assistant tool Eddie samples pixels to decide whether untagged footage is Log, but still takes the specific curve from residual metadata or from the camera brand~\citep{eddie2026log}.
\citet{pearlstein2019retrieval} train a network on about 2.5 million images to distinguish BT.601 from BT.709, a binary classification.
Camera source identification in image forensics distinguishes device models by in-camera traces such as sensor noise~\citep{gloe2010dresden,hosler2019video,cozzolino2019extracting}, not encoding curves: one body can record different curves, and bodies of different brands can record the same curve (\Cref{sec:real}).
LumiGrade~\citep{guo2026lumivideo} evaluates Log video grading but takes the input encoding as given; we examine the step that precedes it.

\noindent\textbf{Benchmarks with ground truth defined by programs or formulas.}
CLEVR~\citep{johnson2017clevr}, DynaMath~\citep{zou2025dynamath} and PRiSM~\citep{imani2025prism} use programs to generate images and verifiable answers, but their test images are synthetic renderings or programmatically drawn charts.
DoRF~\citep{grossberg2003space} collects 201 response curves of film, sensors and digital cameras, from which later work regresses a continuous response function for a single image~\citep{li2017crf,liu2020single}.
In our setting the formulas are the vendor standards that cameras actually execute, the main test object is real footage, and the task is exact discrimination among official curves that lie very close in code value; the same implementation of the formulas also yields the synthetic images, the training-free baseline and the specification reference.

\noindent\textbf{Quantitative visual ability of multimodal models.}
General multimodal benchmarks emphasize high-level semantic understanding and subject reasoning~\citep{goyal2017vqav2,liu2024mmbench,yue2024mmmu}, and ChartQA~\citep{masry2022chartqa} examines reading values from statistical charts.
Closer to ours are CameraBench~\citep{fang2025camerabench}, which estimates the capture parameters of photographs, SNAP~\citep{kotseruba2025snap}, which measures how camera settings affect recognition, PhotoBench~\citep{qi2025photographer}, which tests photographic critique, ColorBench~\citep{liang2025colorbench}, which evaluates color perception and reasoning, and QuantiPhy~\citep{li2026quantiphy}, which shows that estimates of physical quantities in video depend heavily on priors.
Their evidence still rests on visible scene semantics or geometric content, whereas in our task the answer depends on the distribution of low-level pixel code values, and the probe and the re-encoding control separate pixel measurement from guessing from priors (\Cref{tab:compare}).

\begin{table}[t]
  \caption{Comparison with related visual benchmarks. \emph{Answer source}: how the ground truth is obtained. \emph{Synth.}\ / \emph{Real}: program-generated or real test images. \emph{Anti-guess.}: paired perturbations or probes that test guessing from priors. \yes~yes, \partly~partly, \no~no.}
  \label{tab:compare}
  \centering
  \small
  \setlength{\tabcolsep}{3pt}
  \begin{tabularx}{\textwidth}{@{}l>{\raggedright\arraybackslash}X>{\raggedright\arraybackslash}Xccc>{\raggedright\arraybackslash}p{2.9cm}@{}}
    \toprule
    Benchmark & Task & Answer source & Synth. & Real & Anti-guess. & Scale \\
    \midrule
    MMMU~\citep{yue2024mmmu} & College-level QA & Human & \no & \yes & \no & 11.5k questions \\
    ChartQA~\citep{masry2022chartqa} & Chart reading & Human + model & \no & \yes & \no & 32.7k questions \\
    \midrule
    CLEVR~\citep{johnson2017clevr} & Synthetic scene reasoning & Program & \yes & \no & \partly & 100k images, $\sim$1M QA \\
    DynaMath~\citep{zou2025dynamath} & Parametric visual math & Program & \yes & \no & \yes & 501 $\times$ 10 questions \\
    QuantiPhy~\citep{li2026quantiphy} & Physical quantities, video & Program, measured, human & \yes & \yes & \yes & 3.3k questions \\
    \midrule
    CameraBench~\citep{fang2025camerabench} & Photo capture parameters & Camera metadata & \no & \yes & \no & 100 photos \\
    ColorBench~\citep{liang2025colorbench} & Color perception, reasoning & Human + program & \partly & \yes & \yes & 1,448 inst.\ (5.8k+ QA) \\
    LumiGrade~\citep{guo2026lumivideo} & Log video grading & Metrics + user ratings & \no & \yes & \no & 100+ clips \\
    Dresden~\citep{gloe2010dresden} & Camera model forensics & Capture device & \no & \yes & \no & $>$14k images, 73 devices \\
    \midrule
    \bench (ours) & Log curve identification & Records + vendor formulas & \yes & \yes & \yes & 330 real clips + controls \\
    \bottomrule
  \end{tabularx}
\end{table}

\section{CameraLogBench}
\label{sec:bench}

\begin{figure}[t]
  \centering
  \includegraphics[width=\textwidth]{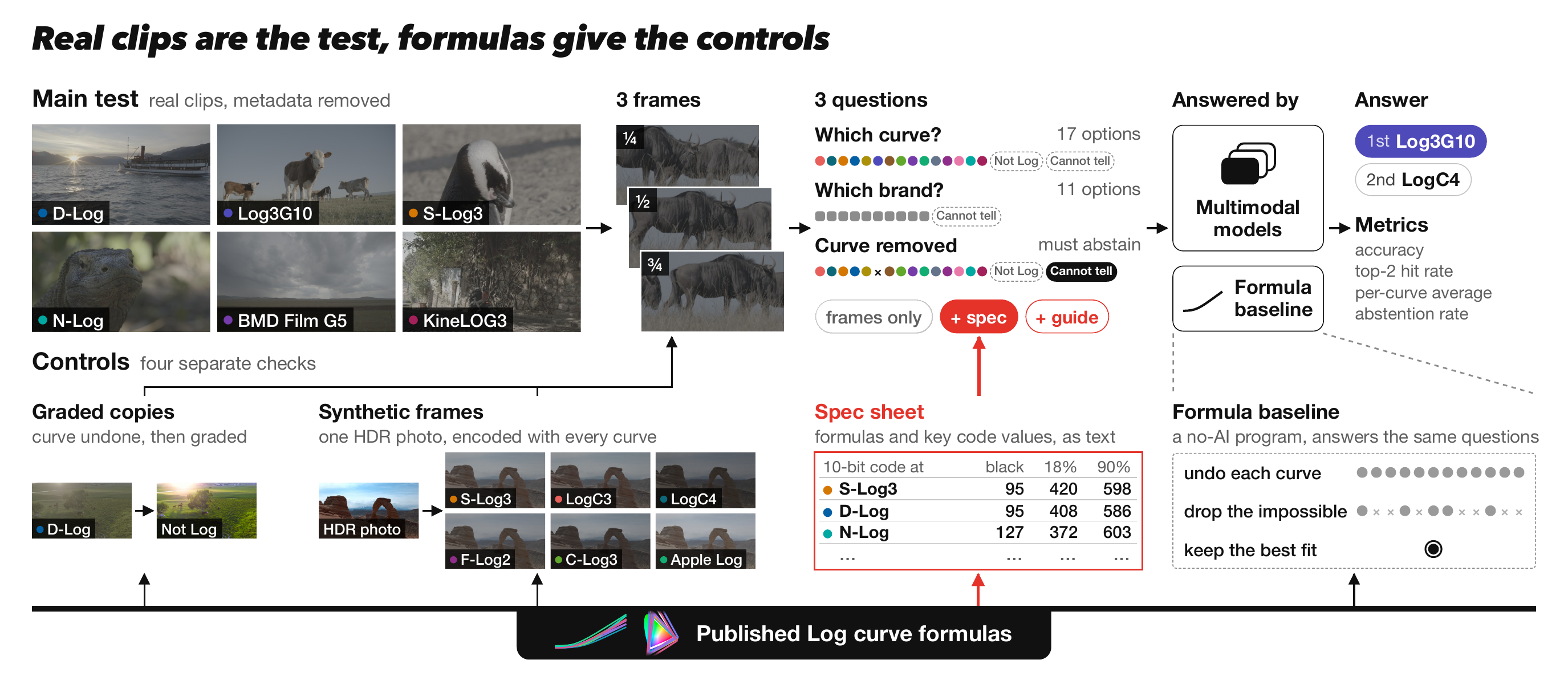}
  \caption{Evaluation pipeline and controls of \bench.
  \textbf{Top row:} three frames of a real clip without metadata are paired with curve identification (17 options), brand identification (11 options) and a probe with the correct option removed (16 options), and given to multimodal models under several prompt tiers; the formula baseline is scored on the same items.
  \textbf{Bottom row:} the four controls derived from the published curve formulas, with arrows marking where each enters the pipeline: graded negatives, synthetic frames, specification digests and the formula baseline (\Cref{sec:controls}).}
  \label{fig:build}
\end{figure}

\bench takes real clips without metadata as its main test and uses four controls derived from vendor formulas to dissect the performance on them (\Cref{fig:build} and \Cref{tab:data}).

\begin{table}[t]
  \caption{Dataset composition and the gamut paired with each curve.
The real set has 330 Log clips and 77 graded negatives.
The camera bodies that recorded the clips are shown in \Cref{fig:stats}a, and the 30 real chart frames are detailed in Appendix~\ref{app:examples} (\Cref{tab:bodies}).
$^{*}$Apple Log 2 and Apple Log share the same transfer curve but are paired with different gamuts; Fujifilm F-Log2 C retains the F-Log2 transfer curve and pairs it with the wider F-Gamut C, so the 12 official curves correspond to the 13 real encodings in the table.}
  \label{tab:data}
  \centering
  \begin{tabular}{@{}llrrrr@{}}
    \toprule
    & & \multicolumn{2}{c}{Real} & \multicolumn{2}{c}{Synthetic} \\
    \cmidrule(lr){3-4}\cmidrule(l){5-6}
    Curve & Gamut & Clips & Charts & Scenes & Charts \\
    \midrule
    Apple Log$^{*}$ & Rec.2020 & 11 & 1 & 50 & 3 \\
    Apple Log 2$^{*}$ & Apple Wide Gamut & 30 & 1 & -- & 3 \\
    ARRI LogC3 & ARRI Wide Gamut 3 & 25 & 1 & 50 & 9 \\
    ARRI LogC4 & ARRI Wide Gamut 4 & 22 & 15 & 50 & 3 \\
    Blackmagic Film Gen5 & Blackmagic Wide Gamut & 16 & 1 & 50 & 3 \\
    Canon C-Log2 & Cinema Gamut & 20 & 2 & -- & 3 \\
    Canon C-Log3 & Cinema Gamut & 21 & 2 & 50 & 3 \\
    DJI D-Log & D-Gamut & 54 & 1 & 50 & 3 \\
    Fujifilm F-Log2 C$^{*}$ & F-Gamut C & 31 & 1 & 50 & 6 \\
    Kinefinity KineLOG3 & Kinefinity Wide Gamut & 5 & -- & 50 & 3 \\
    Nikon N-Log & Rec.2020 & 12 & 2 & 50 & 3 \\
    RED Log3G10 & REDWideGamutRGB & 53 & 1 & 50 & 3 \\
    Sony S-Log3 & S-Gamut3.Cine & 30 & 2 & 50 & 3 \\
    \midrule
    Not Log & -- & 77 & -- & 50 & 2 \\
    \midrule
    \textbf{Total} & & \textbf{407} & \textbf{30} & \textbf{600} & \textbf{50} \\
    \bottomrule
  \end{tabular}
\end{table}

\subsection{Task definition}
\label{sec:task}

The benchmark attaches three questions to every clip.
A real item shows three frames of one clip without metadata, taken at 1/4, 1/2 and 3/4 of its duration and sent together as separate images in temporal order; a synthetic item is a single still frame.
\begin{itemize}[leftmargin=*,nosep]
  \item \textbf{Curve identification, 17 options.}
  The options are 15 ``curve / gamut'' encodings plus ``not a Log encoding'' and ``cannot be determined'' (Appendix~\ref{app:prompts}).
  Fourteen of the encodings are built on the 12 official curves with a published vendor formula (Appendix~\ref{app:curves}); the remaining one, D-Log~M, a common encoding of DJI consumer models, has no published formula and is never the ground truth.
  The model outputs the evidence it observed, its top two candidates and a final choice.
  \item \textbf{Brand identification, 11 options.}
  The options are 10 camera brands plus ``cannot be determined''.
  Since a body of one brand can record the curve of another vendor (\Cref{sec:real}), this question separates device recognition from curve recognition.
  \item \textbf{Probe, 16 options.}
  The frames and the prompt are unchanged and only the correct encoding is removed from the options, so the correct answer is always ``cannot be determined''; a model that switches to another curve exposes guessing from priors.
\end{itemize}

\subsection{Real footage}
\label{sec:real}

\begin{figure}[t]
  \centering
  \includegraphics[width=\textwidth]{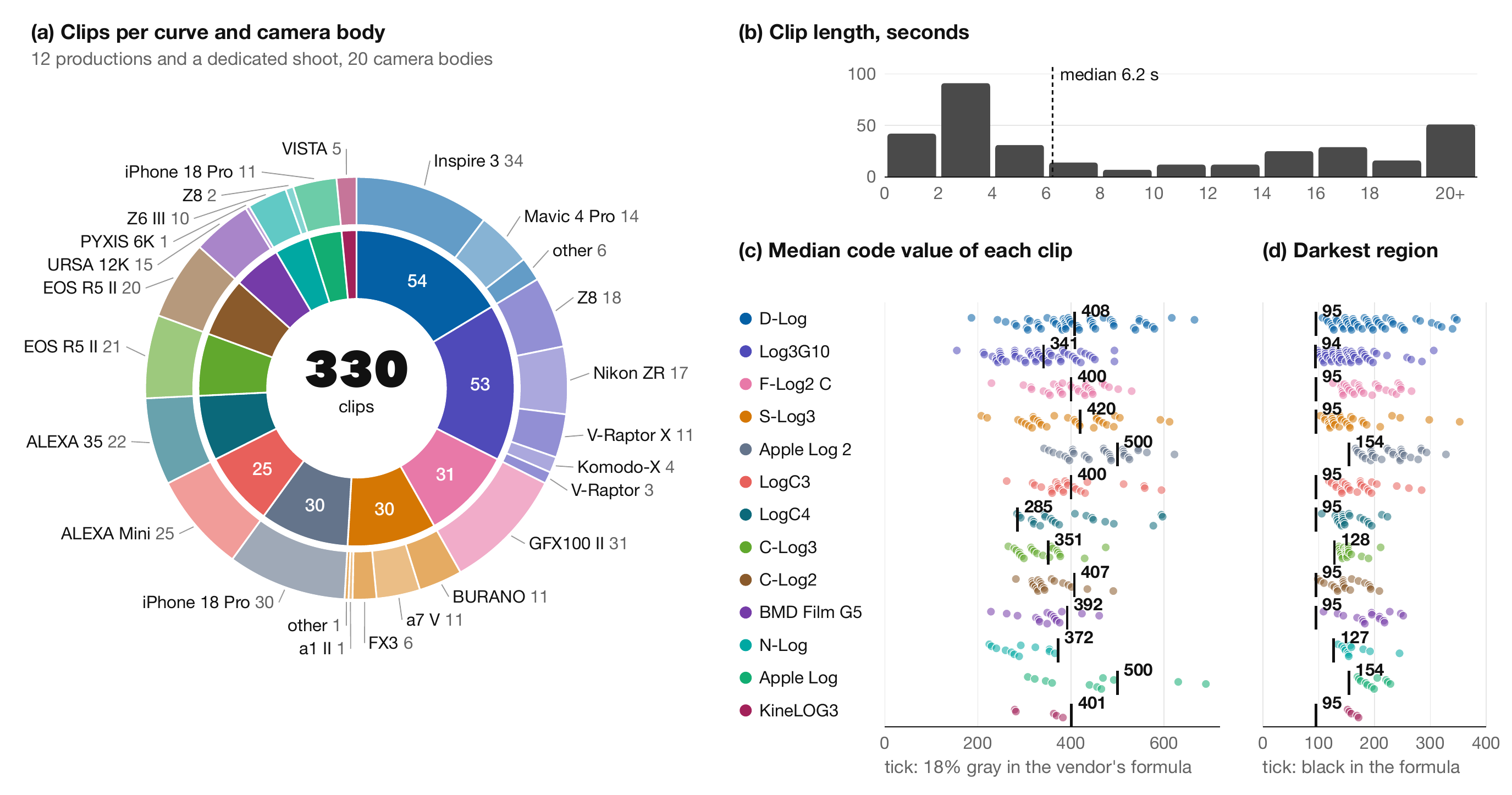}
  \caption{Statistics of the 330 real Log clips. \textbf{(a)} Clips per encoding (inner ring) and per recording body (outer ring). \textbf{(b)} Clip duration (median 6.2\,s). \textbf{(c)} Frame-median code value of each clip (dots) and the 18\% gray level of its curve (vertical line). \textbf{(d)} Darkest region of each clip (dots) and the black level of its curve (vertical line). Code values are 10-bit.}
  \label{fig:stats}
\end{figure}

\noindent\textbf{Collection and label sources.}
The real set contains 330 ungraded Log clips recorded with 20 camera bodies and covers all 12 official curves (\Cref{tab:data}, \Cref{fig:stats}a).
Of these, 170 come from 12 real productions and cover D-Log~\citep{dji2022white}, Log3G10~\citep{red2017white}, S-Log3~\citep{sony2014technical}, N-Log~\citep{nikon2018nlog}, Film Gen5~\citep{blackmagic2021blackmagic} and KineLOG3~\citep{kinefinity2025kinelog3}.
The other 160 come from one dedicated shoot and cover LogC3~\citep{arri2017alexa} and LogC4~\citep{arri2025arri}, C-Log2 and C-Log3~\citep{canon2018canon}, Apple Log~\citep{apple2023apple} and Apple Log~2~\citep{apple2025applelog2}, and F-Log2~C~\citep{fujifilmflog2c}.
All clips are 3840$\times$2160 10-bit ProRes video with durations between 0.67 and 52.4\,s (\Cref{fig:stats}b); the model receives only the extracted frames, which carry no metadata.
Production labels are taken from the project records and cross-checked against the header or the sidecar file of the original recording, and the labels of the dedicated shoot are read from the in-camera settings stored in the original files.

\noindent\textbf{A subset that decouples brand and curve.}
In ordinary shooting brand and curve are tightly bound, but the productions naturally contain a decoupled subset: Nikon bodies record 35 clips of RED Log3G10 internally as R3D, 17 from the ZR and 18 from the Z8, while the Z8 and Z6~III record another 12 clips in Nikon's own N-Log.
This allows a direct test on real footage of whether a model recognizes the curve or the brand.

\noindent\textbf{Code-value distribution.}
In real shooting, framing and exposure vary freely, and a frame cannot be assumed to sit near 18\% gray.
The frame-median code values of the clips scatter widely around the 18\% gray level of their curve and overlap heavily across curves (\Cref{fig:stats}c); for D-Log, whose 18\% gray lies at 408, the medians span 186 to 665.
The darkest region of every clip sits at or---in most clips---well above the curve's black floor (\Cref{fig:stats}d), leaving natural frames without a true-black reference.

\noindent\textbf{Real charts.}
To measure identification when the frame contains a known photometric reference, we additionally shot 30 frames of standard color charts with 10 camera bodies, covering 12 encodings (\Cref{tab:data}).

\subsection{Controls and baselines derived from vendor formulas}
\label{sec:controls}

Accuracy on real footage alone cannot tell whether a model has grasped the photometric regularities or relies on scene preferences.
We implement all 12 published specifications in a single formula code base (Appendix~\ref{app:curves}) and derive from it four controls and two baselines (\Cref{fig:build}, bottom row).

\noindent\textbf{Graded clips.}
To test whether a model can tell unconverted Log footage from restored footage, we take all 5 clips of KineLOG3 and 6 clips of each of the other 12 real encodings, decode them to Rec.709 with the official inverse functions and grade them: 38 with a standard conversion, 26 with a creative look and 13 with lowered contrast that imitates Log footage.
These 77 negatives, whose ground truth is ``not Log'', share their content with the original clips, so the subject matter does not reveal whether a clip has been graded.

\noindent\textbf{Synthetic frames.}
In the real set, different curves are shot with different bodies in different scenes.
To remove this bias we forward-encode linear HDR images, whose pixel values are proportional to scene light, with the official formulas: the mean luminance is aligned to 18\% gray, the image is converted to the gamut paired with the target curve, and the formula maps it to code values.
There are three sets, with examples in Appendix~\ref{app:examples} (\Cref{fig:examples}).
(1)~The \emph{synthetic natural-scene set} takes 100 public HDR scenes: 76 photographs from the HDR Photographic Survey~\citep{fairchild2007hdr} and 24 panoramas from Poly Haven~\citep{polyhaven}; for each of 11 official curves we encode 50 distinct scenes from this pool, and 50 non-Log controls bring the total to 600 frames.
(2)~The \emph{synthetic chart set} contains 50 programmatically drawn frames with a gray card, a 24-patch color chart, a gray ramp, and pure-black and clipped patches, covering all 14 encodings with a published formula.
(3)~The \emph{re-encoding set} inverts 6 real frames and re-encodes each with 11 curves and 2 non-Log versions, 78 frames in total; it keeps real sensor noise and tests whether predictions follow the underlying encoding.

\noindent\textbf{Specification digests.}
To distinguish a lack of parametric knowledge from a lack of measurement ability, we export two tiers of English reference from the same implementation (Appendix~\ref{app:prompts}).
The \emph{spec} tier lists the formulas, the anchor code values and the gamut chromaticity coordinates; the \emph{spec + guide} tier adds a six-step measurement guide that covers the shadow floor, neutral references, highlight clipping plateaus and gamut checks on highly saturated colors.

\noindent\textbf{Two baselines without a large model.}
Two reference baselines quantify the signal carried by the pixels themselves; see \Cref{sec:baselines} for their implementation.
The \emph{formula baseline} is an analytic program with no learned parameters: it assumes each candidate curve in turn, decodes the frame back to scene light and chromaticity, discards the curves that violate physical constraints (shadows below 154, for instance, rule out Apple Log, whose black sits at 154), and selects among the rest the one whose inferred exposure is closest to standard middle gray.
The \emph{small classifier} is a lightweight two-branch convolutional network trained only on formula-synthesized images, which tests whether purely synthetic supervision generalizes to real footage.

\subsection{Formula baseline and small classifier}
\label{sec:baselines}

\noindent\textbf{Formula baseline.}
The formula baseline is an analytic program with no learned parameters that decides on each natural image in three steps.
The first step is statistical measurement: the frame is divided into local blocks of 24$\times$24 pixels whose means smooth single-pixel noise, and the program extracts the code values of the darkest and the brightest block in the frame and detects whether the highlights contain a clipping plateau, formed by many pixels piling up at one code value.
The second step is screening by physical constraints: assuming each candidate curve in turn, the program decodes the blocks back to scene-linear light with its inverse function, converts them to chromaticity with the matrix of the paired gamut, and checks three constraints.
The darkest part of the frame must not lie significantly below the theoretical black level of the curve, no negative light may appear after decoding, and the converted chromaticities must not fall outside the visible spectral locus; a candidate that violates any of the three is discarded.
The third step ranks the surviving candidates by exposure plausibility: the closer the log-average luminance of the decoded frame is to 18\% gray, the higher the score.
In addition, the white papers of LogC3 and D-Log give highlight clipping code values that vary with sensitivity, and an observed highlight plateau that matches one of them earns a bonus.
The candidate with the highest combined score is the first prediction and the runner-up is the second candidate.

On calibrated chart items, the formula baseline uses a version with known patch positions.
The program directly reads the mean code values of the gray card, each step of the gray ramp and 8 standard color patches.
For each candidate curve, it first fits a global exposure offset on the gray series that minimizes the RMS residual between the code values the curve predicts and the measured ones, and records this residual; it then decodes the color patches to chromaticity with the curve and its paired gamut and computes their deviation from the standard chromaticities.
Both errors are independent of the shooting exposure.
A candidate for which either error exceeds about three times its tolerance is discarded, and if all candidates are discarded the program outputs ``not Log''.

For the real charts, the patch positions are located automatically by a ColorChecker detector and checked frame by frame.
Reflecting the physical imaging process of real cameras, the real-capture measurement model has three settings: the reference values of the patches are the measured spectral data published by X-Rite; the gray-ramp fit jointly estimates the shadow lift caused by stray light inside the lens; and the chromaticity tolerance is set to 0.01, the level of the color-reproduction error of real cameras, against 0.005 for the synthetic charts.
This baseline answers 28 of the 30 real chart frames correctly and fails only on D-Log and Log3G10, which have one frame each in the set.

\noindent\textbf{Small classifier.}
The small classifier measures how supervision synthesized purely from the formulas generalizes to real footage.
The model has two branches.
The first is a ResNet-18 trained from scratch, whose input is the tensor of raw code values of the 384$\times$216 frame after area-averaging to half its size, without ImageNet mean and variance normalization so that the absolute code values are preserved.
The second computes 256-bin code-value histograms of all pixels in four channels, R, G, B and luma, and extracts a 128-dimensional feature with a two-layer fully connected network.
The two features are concatenated and fed to a linear classification head that outputs 13 classes: the 12 encodings covered by the synthetic set and ``not Log''.

All training samples are synthesized online.
Each sample is a frame with an ordinary field of view, obtained by perspective projection from a linear HDR panorama with a random heading, pitch and field of view; with a probability of 35\%, a local region of another scene is pasted in to break the association with scene semantics.
Random perturbations of contrast, saturation, exposure and white balance are then applied, simulated highlight clipping, shot and read noise and lens flare are added, and finally a randomly drawn official curve encodes the result with its formula, followed by quantization to 8 bits; the ``not Log'' class instead receives a random color restoration and creative grade.
The training set consists of 729 panoramas from Poly Haven~\citep{polyhaven} and the validation set of 80 panoramas held out by shooting location, and no panorama from the same location as the synthetic test set is used for training.
The network is trained with the AdamW optimizer at a peak learning rate of $10^{-3}$ with a one-cycle schedule and a batch size of 64 for 16,000 steps; on real clips the output probabilities of the three frames are averaged.
The curve-identification accuracy of the classifier falls step by step as the test distribution moves away from the training domain: 0.46 on the held-out validation panoramas, 0.26 on the synthetic natural-scene set (\Cref{tab:charts}) and 0.14 on the 330 real clips (\Cref{tab:real}).
With more training panoramas or four times as many training steps, the accuracy on real footage stays at 0.14--0.15.

\subsection{Evaluation protocol}
\label{sec:protocol}

Each frame is downscaled by area averaging to 1920$\times$1080 and stored as an 8-bit lossless image, with no tone or color-space conversion; Appendix~\ref{app:input} compares it per pixel with the 10-bit original.
Prompts come in four tiers that share the question and the options (Appendix~\ref{app:prompts}, \Cref{tab:tiers}).
\emph{Text only} hides the images and checks whether the wording introduces a bias; \emph{frames only} measures what a model reaches from its internal knowledge; \emph{frames + spec} and \emph{frames + spec + guide} add the two specification digests of \Cref{sec:controls}.
We evaluate real items on all tiers except \emph{frames + spec}, and synthetic items on the three visual tiers; each item is asked once under each of two option orders, and an output with no valid option counts as wrong.

We report four metrics: \emph{accuracy} over all calls; the \emph{top-2 hit rate}, the share of calls whose top two candidates contain the ground truth; the \emph{per-curve average}, the unweighted mean of the per-encoding accuracies, which removes the effect of class sizes; and the \emph{abstention rate}, the share of probe items answered ``cannot be determined''.
Ablations on bit depth, file format and the number of frames are given in Appendix~\ref{app:ablation}.

\section{Experiments}
\label{sec:exp}

\subsection{Setup}
\label{sec:setup}

We evaluate 11 mainstream multimodal models: Claude Opus 5.5~\citep{anthropic2026opus55}, Claude Sonnet 5.5~\citep{anthropic2026sonnet55}, GPT-6.1 Sol~\citep{openai2026gpt61sol}, GPT-6 Astra~\citep{openai2026gpt6astra}, Gemini 3.1 Pro~\citep{google2026gemini31pro}, Gemini 3.8 Flash~\citep{google2026gemini38flash}, Qwen3.8-Max~\citep{alibaba2026qwen38max}, Qwen3.8-Flash~\citep{alibaba2026qwen38flash}, Kimi K3.1 Pro~\citep{moonshot2026kimik31pro}, Kimi K3~\citep{moonshot2026kimik3} and DeepSeek-V4.1-Flash~\citep{deepseek2026v41flash}.
All models receive the same images and prompts, with reasoning effort uniformly set to \texttt{high} and the output limit set to 32k tokens.

We compare against four baselines.
\emph{Random guess} picks uniformly among the options.
\emph{Always D-Log} answers D-Log, the curve with the most real samples, for every item and gives the score obtainable from class imbalance alone.
The \emph{formula baseline} and the \emph{small classifier} of \Cref{sec:controls} never output ``cannot be determined'' and take no part in the probe.

\subsection{Main results on real footage}
\label{sec:main}

\begin{table}[t]
  \caption{Main results on real footage. The first six columns report curve identification on the 330 Log clips under two prompt tiers: accuracy (Acc.), top-2 hit rate (Top-2) and the per-curve average over the 13 real encodings (Avg.). Brand and Not Log use frames only: brand accuracy on the 330 Log clips and the detection rate on the 77 graded negatives. Bold marks the best multimodal model in each column.}
  \label{tab:real}
  \centering
  \small
  \begin{tabular}{@{}lcccccccc@{}}
    \toprule
    & \multicolumn{3}{c}{Frames only} & \multicolumn{3}{c}{Frames + spec + guide} & & \\
    \cmidrule(lr){2-4}\cmidrule(lr){5-7}
    Method / model & Acc. $\uparrow$ & Top-2 $\uparrow$ & Avg. $\uparrow$ & Acc. $\uparrow$ & Top-2 $\uparrow$ & Avg. $\uparrow$ & Brand $\uparrow$ & Not Log $\uparrow$ \\
    \midrule
    Random guess & 0.06 & 0.12 & 0.06 & 0.06 & 0.12 & 0.06 & 0.09 & 0.06 \\
    Always D-Log & 0.16 & -- & 0.08 & 0.16 & -- & 0.08 & 0.16 & -- \\
    Formula baseline & 0.09 & 0.25 & 0.09 & 0.09 & 0.25 & 0.09 & -- & -- \\
    Small classifier & 0.14 & 0.30 & 0.15 & -- & -- & -- & -- & 0.96 \\
    \midrule
    Claude Opus 5.5 & 0.13 & \textbf{0.30} & 0.09 & 0.17 & 0.30 & 0.12 & \textbf{0.22} & 0.53 \\
    Claude Sonnet 5.5 & 0.15 & 0.23 & 0.12 & 0.16 & 0.25 & 0.11 & 0.21 & 0.66 \\
    GPT-6.1 Sol & 0.14 & 0.28 & 0.09 & 0.25 & \textbf{0.41} & \textbf{0.18} & 0.15 & 0.94 \\
    GPT-6 Astra & 0.14 & 0.27 & 0.08 & \textbf{0.27} & \textbf{0.41} & 0.17 & 0.14 & \textbf{0.97} \\
    Gemini 3.1 Pro & 0.19 & 0.27 & \textbf{0.14} & 0.14 & 0.24 & 0.12 & \textbf{0.22} & 0.22 \\
    Gemini 3.8 Flash & \textbf{0.20} & 0.29 & \textbf{0.14} & 0.20 & 0.31 & 0.13 & 0.18 & 0.49 \\
    Qwen3.8-Max & 0.06 & 0.19 & 0.05 & 0.05 & 0.19 & 0.07 & 0.20 & 0.87 \\
    Qwen3.8-Flash & 0.14 & 0.24 & 0.10 & 0.16 & 0.28 & 0.12 & 0.19 & 0.38 \\
    Kimi K3.1 Pro & \textbf{0.20} & 0.28 & 0.13 & 0.20 & 0.26 & 0.14 & 0.19 & 0.61 \\
    Kimi K3 & 0.18 & 0.27 & 0.12 & 0.16 & 0.25 & 0.12 & 0.21 & 0.42 \\
    DeepSeek-V4.1-Flash & 0.15 & 0.25 & 0.10 & 0.18 & 0.31 & 0.14 & 0.20 & 0.60 \\
    \bottomrule
  \end{tabular}
\end{table}

\Cref{tab:real} shows that current models are still far from usable on real footage.
In the text-only control, every model chooses ``cannot be determined'' in 99--100\% of its calls, so the question and the options offer no shortcut.
With frames only, the curve accuracy of the 11 models lies between 0.06 and 0.20.
Always answering D-Log, which accounts for 54 of the 330 clips, scores 0.16; the best models, Gemini 3.8 Flash and Kimi K3.1 Pro, exceed it by only 4 percentage points, and 7 of the 11 models fall below it.
Under the per-curve average, always answering D-Log scores $1/13 \approx 0.08$ and the models score 0.05 to 0.14.
Brand identification likewise stays near the frequency baseline: the models score 0.14 to 0.22, while always answering DJI scores 0.16.
With the specification digest and the measurement guide, only GPT-6 Astra (0.27) and GPT-6.1 Sol (0.25) improve markedly, by 13 and 11 percentage points; the other 9 models change by between $-0.05$ and $+0.04$.

\noindent\textbf{Answer preferences and abstention.}
The output distributions (Appendix~\ref{app:answers}, \Cref{tab:dist}) reveal strong frequency priors.
With frames only, 10 of the 11 models predict S-Log3 most often, in 27--60\% of their outputs, although this curve makes up only 9\% of the real set; only Claude Sonnet 5.5 predicts ARRI LogC3 most often, in 39\% of its outputs.
GPT-6 Astra and GPT-6.1 Sol instead choose ``cannot be determined'' in 46\% and 32\% of their calls; counting only non-abstaining calls, their conditional accuracy reaches 0.26 and 0.21 (Appendix~\ref{app:answers}, \Cref{tab:dist}), above the best overall accuracy of 0.20.
With the specification and the guide, their abstention drops to 6--7\%, and they open a clear lead on less common curves such as Log3G10 and C-Log3.

\noindent\textbf{Per-encoding coverage and comparison with the baselines.}
Correct predictions concentrate on a few common curves such as D-Log, S-Log3 and Apple Log, and accuracy on most other encodings is close to zero (Appendix~\ref{app:answers}, \Cref{tab:percurve}); the best scores on LogC3 and LogC4 are 0.32 and 0.14.
On the other 8 encodings, 188 clips in total, no model exceeds 0.11 on any single encoding, and none exceeds 0.03 on Apple Log~2 and F-Log2~C; on the 30 Apple Log~2 clips, the 11 models produce only 1 correct answer in total.
With the specification and the guide, the two GPT models jump to 0.45 and 0.34 on Log3G10, where no other model exceeds 0.10, and also rise to 0.31 and 0.40 on C-Log3.
The 77 graded negatives, in contrast, are detected reliably by strong models (GPT-6 Astra 0.97, GPT-6.1 Sol 0.94) and by the small classifier (0.96).
The two baselines without a large model are equally hampered by the exposure and imaging variation of real scenes.
The formula baseline takes a frame-average luminance at 18\% gray as its default exposure; it reaches 0.71 on the synthetic natural-scene set generated under this setting (\Cref{sec:charts}), but 0.09 on the real set, where exposure varies freely with the scene.
The small classifier reaches only 0.14 on real footage: synthetic supervision teaches a network to tell whether footage has been graded, but not to tell curves apart across real noise and unknown exposure.

\subsection{Why models fail}
\label{sec:why}

\begin{table}[t]
  \caption{Probe and re-encoding diagnostics. \emph{Abstention}: abstention rate on the probe items of the 330 real clips. \emph{Switch}: share of the calls answered correctly on the regular curve question that switch to an incorrect curve on the probe. \emph{Re-encoding}: accuracy and same-answer share on the 66 re-encoded Log images. Switch and Re-encoding use frames only; + guide denotes the specification with the measurement guide. Bold marks the best multimodal model.}
  \label{tab:probe}
  \centering
  \footnotesize
  \setlength{\tabcolsep}{3pt}
  \begin{tabular}{@{}lccccc@{}}
    \toprule
    & \multicolumn{2}{c}{Abstention $\uparrow$} & & \multicolumn{2}{c}{Re-encoding} \\
    \cmidrule(lr){2-3}\cmidrule(l){5-6}
    Method / model & Frames & + guide & Switch $\downarrow$ & Acc. $\uparrow$ & Same $\downarrow$ \\
    \midrule
    Formula baseline & -- & -- & -- & 0.17 & 0.52 \\
    Small classifier & -- & -- & -- & 0.45 & 0.36 \\
    \midrule
    Claude Opus 5.5 & 0.00 & 0.00 & 100\% & 0.05 & 0.74 \\
    Claude Sonnet 5.5 & 0.00 & 0.00 & 100\% & \textbf{0.08} & 0.68 \\
    GPT-6.1 Sol & 0.33 & 0.07 & 83\% & 0.03 & 0.80 \\
    GPT-6 Astra & \textbf{0.46} & 0.12 & \textbf{80\%} & 0.02 & 0.89 \\
    Gemini 3.1 Pro & 0.00 & 0.00 & 100\% & 0.06 & 0.67 \\
    Gemini 3.8 Flash & 0.00 & 0.00 & 100\% & \textbf{0.08} & 0.73 \\
    Qwen3.8-Max & 0.00 & 0.00 & 95\% & 0.03 & 0.61 \\
    Qwen3.8-Flash & 0.03 & \textbf{0.17} & 97\% & 0.07 & \textbf{0.55} \\
    Kimi K3.1 Pro & 0.00 & 0.00 & 100\% & \textbf{0.08} & 0.68 \\
    Kimi K3 & 0.00 & 0.00 & 99\% & 0.07 & 0.64 \\
    DeepSeek-V4.1-Flash & 0.01 & 0.01 & 100\% & \textbf{0.08} & 0.58 \\
    \bottomrule
  \end{tabular}
\end{table}

A correct answer on a real item may come from photometric measurement of the pixel code values, from guessing the brand from scene features such as an aerial viewpoint, or from falling back on a frequent encoding name; four diagnostics separate these.

\noindent\textbf{Probe.}
Once the correct curve is removed from the options, a model that answers by verifying pixels should choose ``cannot be determined''.
With frames only, however, the abstention rate of every model except GPT-6 Astra (0.46) and GPT-6.1 Sol (0.33) is at most 0.03 (\Cref{tab:probe}).
With the specification, the abstention rate of the GPT models falls to 0.07--0.12, and the highest, Qwen3.8-Flash, reaches only 0.17.
Even on items answered correctly in the regular curve question, the GPT models switch to an incorrect curve on 80--83\% of the probes, and the other 9 models on 95--100\%; for all 11 models the most frequent destination is D-Log~M, of the same brand as D-Log.
Most of the earlier correct answers were therefore prior guesses made by association with the scene and the brand.

\noindent\textbf{Similar encodings of the same brand.}
The per-brand statistics (Appendix~\ref{app:answers}, \Cref{tab:brand}) reveal two mechanisms.
The first is confusing the encoding after recognizing the brand: on the 54 real D-Log clips with frames only, the 11 models combined recognize the DJI brand at a rate of 0.71, yet 15\% of their answers select the higher-contrast D-Log~M, with Qwen3.8-Max, Claude Opus 5.5 and Qwen3.8-Flash reaching 59\%, 35\% and 30\%, respectively.
The second is drifting toward a more familiar, older encoding when the brand is not recognized: apart from DJI (0.71) and Sony (0.58), the pooled recognition rate of each of the other 8 brands is at most 0.07.
On Apple Log~2, where the brand recognition rate is only 0.02, the models are almost never correct, yet 17\% of their answers select the older Apple Log, with Claude Sonnet 5.5 reaching 43\%; on LogC4, 8\% likewise select the older LogC3, with Claude Sonnet 5.5 and Kimi K3 at 32\% and 23\%, respectively.

\noindent\textbf{The decoupled brand--curve subset.}
On the 35 real RED Log3G10 clips shot on Nikon bodies (Appendix~\ref{app:answers}, \Cref{tab:brand}), the pooled curve accuracy of the 11 models is only 0.03 and the brand accuracy 0.01, with the highest single-model curve accuracy at 0.13 for Claude Opus 5.5; predictions fall back on the common S-Log3 and LogC3, at 35\% and 32\%, respectively.
Once strong scene cues such as aerial footage are gone, the models revert to the frequency prior over encodings.

\noindent\textbf{Re-encoding.}
Inverting one real frame and re-encoding it with the 11 official curves yields 11 versions with identical content and different ground truths.
The \emph{same-answer share} is the share of these 11 versions taken by the most frequent prediction: $1/11 \approx 0.09$ when all are answered correctly and 1 when the model repeats the same answer across all encodings, so a lower value indicates that the predictions follow the pixel code values.
The formula baseline and the small classifier reach accuracies of 0.17 and 0.45 with same-answer shares of 0.52 and 0.36 (\Cref{tab:probe}).
The 11 multimodal models reach only 0.02--0.08 with shares of 0.55--0.89; the high share of the GPT models mainly stems from their answering ``cannot be determined'' in about 70--80\% of the calls.

\subsection{The role of gray cards and specification digests}
\label{sec:charts}

\begin{table}[t]
  \caption{Effect of a photometric reference and of the specification digest.
Curve-identification accuracy on the 50 synthetic charts, the 30 real charts and the 550 synthetic natural-scene Log images, 50 per curve, under three prompt tiers; ``+ guide'' denotes the specification plus the measurement guide.
Bold marks the best multimodal model in each column.}
  \label{tab:charts}
  \centering
  \small
  \setlength{\tabcolsep}{2.6pt}
  \begin{tabular}{@{}lccccccccc@{}}
    \toprule
    & \multicolumn{3}{c}{Synthetic charts $\uparrow$} & \multicolumn{3}{c}{Real charts, per-curve avg.\ $\uparrow$} & \multicolumn{3}{c}{Synthetic natural scenes $\uparrow$} \\
    \cmidrule(lr){2-4}\cmidrule(lr){5-7}\cmidrule(l){8-10}
    Method / model & Frames & + spec & + guide & Frames & + spec & + guide & Frames & + spec & + guide \\
    \midrule
    Formula baseline, known patches & \multicolumn{3}{c}{1.00} & \multicolumn{3}{c}{0.83} & \multicolumn{3}{c}{--} \\
    Formula baseline, blind & \multicolumn{3}{c}{--} & \multicolumn{3}{c}{--} & \multicolumn{3}{c}{0.71} \\
    Small classifier & \multicolumn{3}{c}{0.08} & \multicolumn{3}{c}{--} & \multicolumn{3}{c}{0.26} \\
    \midrule
    Claude Opus 5.5 & 0.42 & 0.47 & 0.42 & 0.04 & 0.13 & 0.13 & 0.11 & 0.13 & 0.12 \\
    Claude Sonnet 5.5 & 0.30 & 0.44 & 0.28 & 0.04 & 0.23 & 0.13 & 0.09 & 0.09 & 0.09 \\
    GPT-6.1 Sol & \textbf{0.46} & 0.68 & 0.67 & 0.22 & 0.43 & 0.37 & 0.04 & \textbf{0.16} & \textbf{0.19} \\
    GPT-6 Astra & 0.39 & \textbf{0.75} & \textbf{0.82} & \textbf{0.30} & \textbf{0.46} & \textbf{0.56} & 0.03 & 0.15 & 0.17 \\
    Gemini 3.1 Pro & 0.14 & 0.19 & 0.21 & 0.21 & 0.11 & 0.08 & \textbf{0.12} & 0.10 & 0.11 \\
    Gemini 3.8 Flash & 0.19 & 0.19 & 0.19 & 0.22 & 0.08 & 0.16 & \textbf{0.12} & 0.13 & 0.13 \\
    Qwen3.8-Max & 0.05 & 0.05 & 0.05 & 0.02 & 0.00 & 0.00 & 0.06 & 0.05 & 0.06 \\
    Qwen3.8-Flash & 0.09 & 0.15 & 0.18 & 0.03 & 0.13 & 0.14 & 0.08 & 0.11 & 0.08 \\
    Kimi K3.1 Pro & 0.22 & 0.22 & 0.27 & 0.17 & 0.11 & 0.06 & 0.11 & 0.13 & 0.13 \\
    Kimi K3 & 0.15 & 0.14 & 0.19 & 0.08 & 0.07 & 0.02 & 0.11 & 0.10 & 0.09 \\
    DeepSeek-V4.1-Flash & 0.08 & 0.12 & 0.10 & 0.05 & 0.19 & 0.16 & 0.10 & 0.11 & 0.11 \\
    \bottomrule
  \end{tabular}
\end{table}

\Cref{tab:charts} disentangles two bottlenecks: lacking a photometric reference in the frame versus failing to apply the written specification.

\noindent\textbf{With a calibrated chart in the frame, the specification digest helps substantially.}
On the synthetic chart set, the formula baseline that reads patches at known positions reaches 1.00.
With frames only, the best models are GPT-6.1 Sol (0.46), Claude Opus 5.5 (0.42) and GPT-6 Astra (0.39).
With the specification digest, GPT-6 Astra jumps to 0.75--0.82 and GPT-6.1 Sol to 0.67--0.68, and the two Claude models reach 0.44--0.47 with the specification alone.
Frontier models can thus estimate the code values of the patches and reason against the parameter table.

\noindent\textbf{On real charts, only the GPT models benefit markedly from the specification.}
Of the 30 real chart frames, LogC4 has 15, one per sensitivity setting, and the other encodings 1--2 each, so the real-chart columns report the per-curve average.
Despite the color-reproduction error of real cameras, on-set lighting and sensor noise, the formula baseline that reads patches at known positions answers 28 of the 30 frames correctly, a per-curve average of 0.83: as long as the frame contains a reference of known reflectance and chromaticity, the pixels themselves suffice to discriminate most curves.
With frames only, the best model, GPT-6 Astra, scores 0.30, and GPT-6.1 Sol, Gemini 3.8 Flash and Gemini 3.1 Pro score 0.21--0.22.
With the specification digest, GPT-6 Astra improves to 0.46--0.56 and GPT-6.1 Sol to 0.37--0.43.
None of the other 9 models exceeds 0.23, and Gemini 3.1 Pro and Kimi K3.1 Pro drop from 0.21 and 0.17 to 0.06--0.11 once the specification is introduced.

\noindent\textbf{Natural images without a reference show a pronounced specification--measurement gap.}
In the synthetic natural-scene set, the frames for each curve are drawn from the same 100 scenes, so image semantics offer no shortcut; the set is generated with the frame-average luminance aligned to 18\% gray, and the blind formula baseline, which takes this as its default exposure, reaches 0.71.
All multimodal models, however, score 0.03--0.12 with frames only, close to random guessing (0.06).
With the specification digest, GPT-6.1 Sol rises to 0.16--0.19 and GPT-6 Astra to 0.15--0.17, and no other model gains more than 0.03, even though the measurement guide spells out checks that need no gray card, such as the shadow floor, highlight clipping plateaus and saturated-gamut checks.
A written specification thus cannot replace a photometric anchor in the frame: understanding the specification does not yet translate into reliable pixel measurement.

\section{Conclusion}
\label{sec:conclusion}

As the first publicly released subset of MediaStormBench, \bench examines whether multimodal models can identify a camera's Log encoding from frames without metadata, combining 330 professionally shot clips with controls and baselines derived from the published vendor formulas.
Across 11 mainstream models, identification on real footage is far from solved: the best accuracy is only 0.20 with frames only and rises to 0.27 with the specification and the measurement guide, while always answering D-Log, the most frequent curve, already yields 0.16.
Correct answers stem mainly from scene and brand priors: once the correct option is removed, the abstention rate of 9 models is at most 0.03, and 80--100\% of the previously correct answers switch to an incorrect curve.
The vendor specification lifts the best accuracy on synthetic charts from 0.46 to 0.82, but no model exceeds 0.19 on synthetic natural scenes without a reference.
What current models lack is thus not knowledge of vendor curve parameters but the ability to measure pixel code values quantitatively; \bench isolates this ability from high-level semantic understanding and offers a clear yardstick for low-level quantitative vision in foundation models.

\noindent\textbf{Limitations and future directions.}
First, in data and input format, the benchmark covers ungraded Log footage and uses 8-bit lossless frames, the standard input of multimodal model APIs; for 6 of the official curves, the real footage comes from a single dedicated shoot with a single camera body each.
Subsequent work can extend the real footage to more productions and bodies, support native 10-bit input, and recover the original encoding of graded footage.
More important is the shift from testing models on their own to evaluating them with tools in real workflows: all models in this paper are evaluated without any external tools, to establish the perceptual baseline of the models themselves, whereas in real post-production colorists rely on professional tools for this judgment.
Future work can bring tool calling and an agent harness into the evaluation---for example, letting models read pixel values with an eyedropper, generate waveforms and vectorscopes, write scripts that screen candidates against physical constraints, and even apply candidate color-space transforms in software and check the converted image---to test whether models can reliably handle the task in interactive workflows close to real production pipelines.
Along this line, later subsets of MediaStormBench will continue to draw on other real problems across the whole film and video production workflow.

\bibliography{main}

\clearpage
\appendix
\section{Key values of the curves and vendor documents}
\label{app:curves}

Code values are on the 10-bit scale (0--1023) and are all computed from the mathematical formulas published in the vendors' white papers; the slope per stop is the increase in code value for each doubling of scene light around an 18\% gray card.

\begin{center}
  \small
  \setlength{\tabcolsep}{3.5pt}
  \begin{tabular}{@{}lccclc@{}}
    \toprule
    Curve & Black & 18\% gray & Slope (CV per stop) & Paired gamut & Vendor document \\
    \midrule
    ARRI LogC3 & 95 & 400 & 72 & ARRI Wide Gamut 3 & \citealp{arri2017alexa} \\
    ARRI LogC4 & 95 & 285 & 57 & ARRI Wide Gamut 4 & \citealp{arri2025arri} \\
    Sony S-Log3 & 95 & 420 & 75 & S-Gamut3.Cine & \citealp{sony2014technical} \\
    DJI D-Log & 95 & 408 & 74 & D-Gamut & \citealp{dji2022white} \\
    RED Log3G10 & 94 & 341 & 63 & REDWideGamutRGB & \citealp{red2017white} \\
    Canon C-Log2 & 95 & 407 & 70 & Cinema Gamut & \citealp{canon2018canon} \\
    Canon C-Log3 & 128 & 351 & 85 & Cinema Gamut & \citealp{canon2018canon} \\
    Blackmagic Film Gen5 & 95 & 392 & 60 & Blackmagic Wide Gamut & \citealp{blackmagic2021blackmagic} \\
    Apple Log & 154 & 500 & 83 & Rec.2020 & \citealp{apple2023apple} \\
    Fujifilm F-Log2 & 95 & 400 & 71 & F-Gamut & \citealp{fujifilmflog2} \\
    Nikon N-Log & 127 & 372 & 83 & Rec.2020 & \citealp{nikon2018nlog} \\
    Kinefinity KineLOG3 & 95 & 401 & 76 & Kinefinity Wide Gamut & \citealp{kinefinity2025kinelog3} \\
    \bottomrule
  \end{tabular}
\end{center}

Among the 15 encoding options of the curve question, Apple Log~2~\citep{apple2025applelog2} and Fujifilm F-Log2~C~\citep{fujifilmflog2c} share the transfer curve of Apple Log and F-Log2, respectively, and pair it with a wider gamut (Apple Wide Gamut and F-Gamut~C), yielding 14 encodings with a published formula.
The remaining option, DJI D-Log~M, has no published formula, so it is not listed in the table and the benchmark has no item with it as ground truth (\Cref{sec:task}).

\section{Input frames versus the 10-bit originals}
\label{app:input}

On all 330 real clips, we convert the 3840$\times$2160 10-bit original video frames and the 1920$\times$1080 8-bit lossless frames given to the models to the 10-bit code-value scale and compare them frame by frame, pooling the R, G and B channels:

\begin{center}
  \small
  \setlength{\tabcolsep}{4pt}
  \begin{tabular}{@{}lcc@{}}
    \toprule
    Statistic & Mean over 330 clips & Max.\ deviation over 330 clips \\
    \midrule
    Change of the frame mean & $-0.5$ & 1.3 \\
    Change of the frame median & $-0.5$ & 3.7 \\
    Change of the darkest 1\% & $-0.3$ & 6.0 \\
    Change of the brightest 1\% & 0.0 & 6.8 \\
    Per-pixel error against the unrounded downscaled frame & 1.4 (RMS) & 4.0 \\
    \bottomrule
  \end{tabular}
\end{center}

The losslessly stored images are identical, pixel by pixel, to the uncompressed 8-bit images on all 330 clips.
The per-pixel RMS error does not exceed one 8-bit quantization step, that is, 4 code values on the 10-bit scale, and the shifts at the darkest and brightest 1\% percentiles come from the 2$\times$2 area-average downsampling, which smooths isolated extreme noisy pixels.

\section{Data examples}
\label{app:examples}

In the real chart set, every camera body shoots one frame at a commonly used base sensitivity.
The Sony BURANO, the Canon EOS R5 Mark~II and the Nikon ZR add one high-sensitivity setting, and the ARRI ALEXA~35 covers all 15 settings from EI~200 to EI~6400 (\Cref{tab:bodies}).
Changing the sensitivity alters the noise level of the image and the highlight clipping point of the sensor, but, except for LogC3, the mathematical definition of the transfer curve itself does not change with sensitivity; charts of one curve shot across sensitivities can therefore be used to test the stability of identification.
The formula baseline that reads patches at known positions is correct at every sensitivity setting of these four bodies---all 15 EI settings of the ALEXA~35 and both ISO settings on the BURANO, EOS R5 Mark~II (both curves) and ZR---so its decision does not fluctuate with sensitivity.

\begin{table}[!htbp]
  \caption{Camera bodies and sensitivity settings of the real chart set: 10 bodies and 30 frames, one frame per sensitivity setting.}
  \label{tab:bodies}
  \centering
  \setlength{\tabcolsep}{3pt}
  \begin{tabular}{@{}lllc@{}}
    \toprule
    Camera body & Curve & Sensitivity & Frames \\
    \midrule
    ARRI ALEXA 35 & LogC4 & EI 200--6400, 15 settings & 15 \\
    ARRI ALEXA Mini & LogC3 & EI 800 & 1 \\
    Sony BURANO & S-Log3 & ISO 800, 3200 & 2 \\
    Canon EOS R5 Mark II & C-Log2, C-Log3 & ISO 800, 4000 & 4 \\
    RED V-RAPTOR 8K VV & Log3G10 & ISO 800 & 1 \\
    Nikon ZR & N-Log & ISO 800, 6400 & 2 \\
    Blackmagic URSA Cine 17K & Film Gen5 & ISO 800 & 1 \\
    DJI Ronin 4D & D-Log & ISO 800 & 1 \\
    Apple iPhone 18 Pro & Apple Log, Apple Log 2 & ISO 200 & 2 \\
    Fujifilm GFX100 II & F-Log2 C & ISO 800 & 1 \\
    \bottomrule
  \end{tabular}
\end{table}

\begin{figure}[!htbp]
  \centering
  \includegraphics[width=\textwidth]{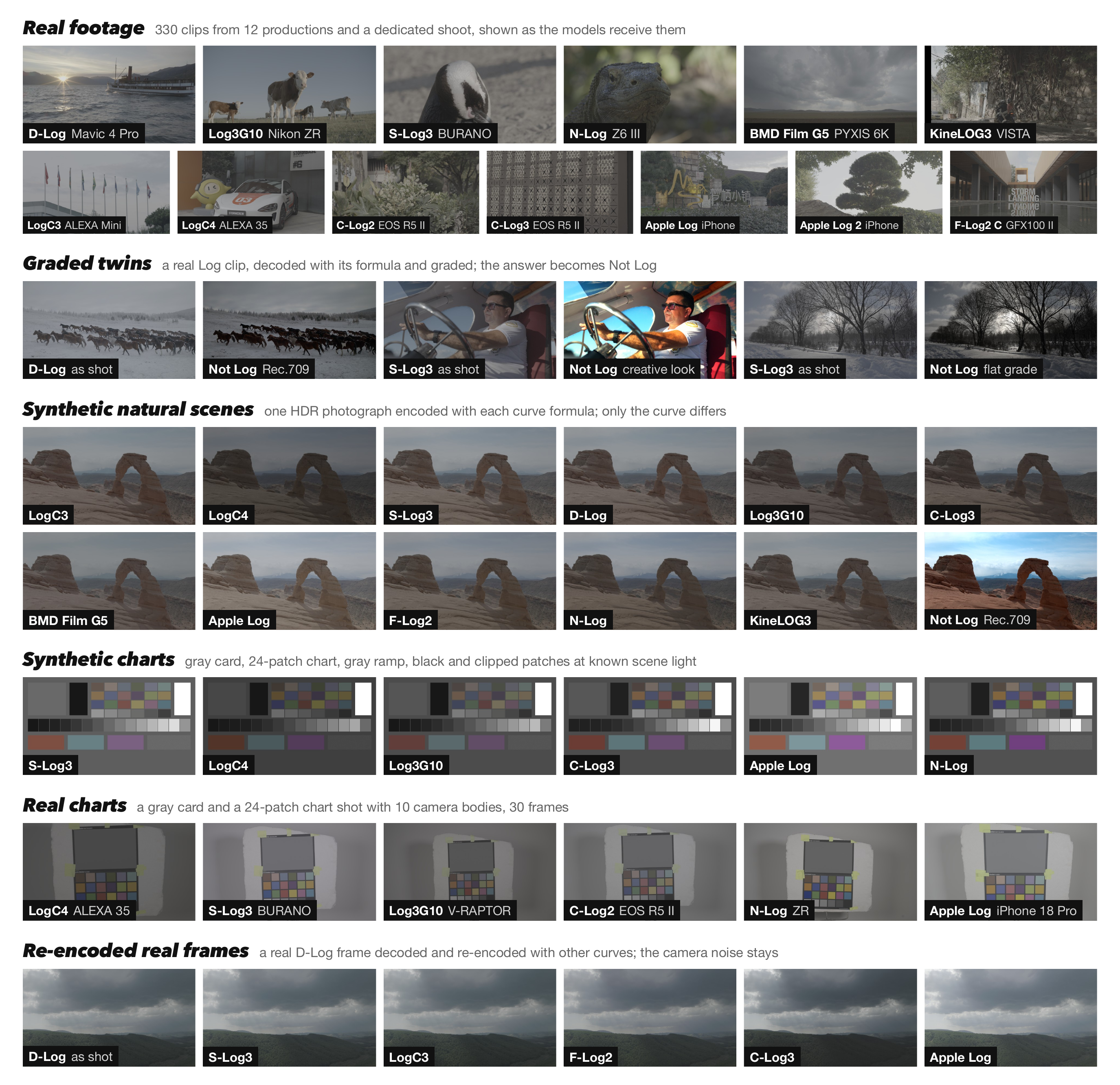}
  \caption{Examples of input frames for each kind of data in the benchmark, all displayed directly with the 8-bit lossless original code values given to the models.
  \textbf{Rows 1--2 (real clips):} one representative clip for each of the 13 real encodings, row 1 from real productions and row 2 from the dedicated shoot.
  \textbf{Row 3 (graded negatives):} three original real clips and their standard Rec.709 conversion, creative look and low-contrast imitation-Log versions, whose ground truth is always ``not Log''.
  \textbf{Rows 4--5 (synthetic natural scenes):} one HDR photograph encoded with the 11 official formulas, with a Rec.709 non-Log control in the last cell.
  \textbf{Row 6 (synthetic charts):} program-generated calibration charts with a gray card, a 24-patch color chart, a gray ramp, and pure-black and clipped patches.
  \textbf{Row 7 (real charts):} physical gray cards and 24-patch color charts shot with different camera bodies under on-set lighting.
  \textbf{Row 8 (re-encoding):} one real D-Log frame inverted and re-encoded with other candidate curves, keeping real sensor noise.}
  \label{fig:examples}
\end{figure}

\clearpage
\section{A complete worked example}
\label{app:worked}

This section shows the input frames and the complete reasoning output of a model on one real curve-identification item.
The clip was shot with a DJI Inspire~3 and its ground truth is DJI D-Log.
The model tested is Claude Opus 5.5.
With frames only, it recognizes DJI equipment from the aerial viewpoint but wrongly selects D-Log~M of the same brand.
In the frames + spec + guide tier shown below, it combines the code value of the shadow floor with the contrast slope of the midtones to rule out D-Log~M and selects the correct option~D under both option orders.

\begin{figure}[H]
  \centering
  \includegraphics[width=\textwidth]{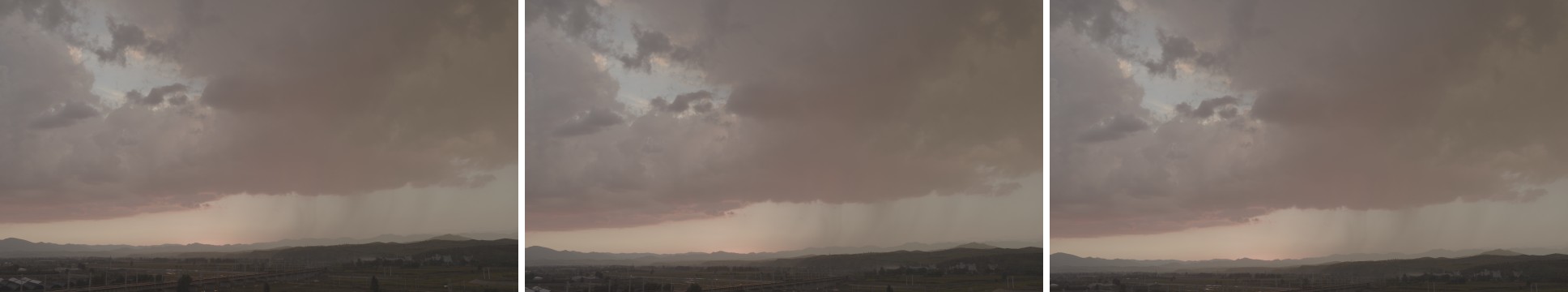}
  \caption*{\textbf{Input frames.} Three 8-bit lossless frames taken at 25\%, 50\% and 75\% of the clip duration.}
\end{figure}

\noindent\textbf{Model output, verbatim.}
The prompt template is given in Appendix~\ref{app:prompts}; under this option order, option~\texttt{D} is DJI D-Log and \texttt{E} is D-Log~M, and the model finally chooses \texttt{D} with \texttt{E} as its second choice.
\begin{lstlisting}
{
 "open_answer": "DJI drone footage in a flat log profile, most likely DJI D-Log / D-Gamut, with D-Log M as the close alternative.",
 "evidence": [
  "The content points to a DJI drone: an aerial view over a valley with a railway viaduct, with a slow camera move across the three frames.",
  "The image is very flat and desaturated, with no display contrast curve. The pink sunset and blue sky patch show only weak saturation, which suggests a log encoding plus a wide gamut rather than Rec.709.",
  "The darkest foreground ground and trees average about 8-bit 35-45, roughly 140-180 on the 10-bit scale. This sits above the 95 black family with no true black anywhere in frame. That is consistent with D-Log, while D-Log M would let dusk shadows fall lower.",
  "The large mid-tone cloud mass sits around 8-bit 120-130 (about 480-520 10-bit). That is near or slightly above the D-Log grey of 408 for a bright overcast sky.",
  "The brightest areas (horizon glow and the blue sky patch) reach about 8-bit 190-200 (about 760-800 10-bit), with no hard clip plateau. This fits below the D-Log clip of about 948.",
  "The tonal range of roughly 160 to 800 is compressed with low contrast per stop, which favours true D-Log over the higher-contrast D-Log M."
 ],
 "top2": [
  "D",
  "E"
 ],
 "choice": "D"
}
\end{lstlisting}

\section{Answer distributions and details of the prior shortcuts}
\label{app:answers}

The following three tables give per-model details for \Cref{sec:main} and \Cref{sec:why}: each model's output distribution on the real-footage curve question; per-encoding recall; and per-brand recognition rates, confusion rates between similar encodings of the same brand, and results on the decoupled brand--curve subset.

\begin{table}[H]
  \caption{Output distribution on the real-footage curve question with frames only, over the 330 Log clips; each row sums to 1. ``Other real'' aggregates the other 11 encodings of the real set; ``not in real set'' is F-Log2, which has no real clips; ``cannot tell'' is the option ``cannot be determined''. The last column is the conditional accuracy over the calls that give a specific prediction, after excluding ``cannot be determined''. Bold marks the best value.}
  \label{tab:dist}
  \centering
  \small
  \setlength{\tabcolsep}{3pt}
  \begin{tabular}{@{}lcccccccc@{}}
    \toprule
    & \multicolumn{7}{c}{Share of each kind of answer} & Accuracy when \\
    \cmidrule(lr){2-8}
    Model & D-Log & S-Log3 & Other real & Not in real set & D-Log M & Not Log & Cannot tell & answering \\
    \midrule
    Claude Opus 5.5 & 0.06 & 0.48 & 0.35 & 0.01 & 0.09 & 0.01 & 0.00 & 0.13 \\
    Claude Sonnet 5.5 & 0.14 & 0.08 & 0.69 & 0.06 & 0.02 & 0.01 & 0.00 & 0.15 \\
    GPT-6.1 Sol & 0.11 & 0.36 & 0.18 & 0.00 & 0.03 & 0.00 & 0.32 & 0.21 \\
    GPT-6 Astra & 0.11 & 0.27 & 0.13 & 0.00 & 0.03 & 0.00 & 0.46 & \textbf{0.26} \\
    Gemini 3.1 Pro & 0.25 & 0.35 & 0.38 & 0.02 & 0.00 & 0.00 & 0.00 & 0.19 \\
    Gemini 3.8 Flash & 0.17 & 0.45 & 0.30 & 0.04 & 0.03 & 0.01 & 0.00 & 0.20 \\
    Qwen3.8-Max & 0.01 & 0.39 & 0.20 & 0.03 & 0.22 & 0.16 & 0.00 & 0.06 \\
    Qwen3.8-Flash & 0.15 & 0.60 & 0.12 & 0.01 & 0.10 & 0.01 & 0.02 & 0.14 \\
    Kimi K3.1 Pro & 0.15 & 0.54 & 0.25 & 0.01 & 0.04 & 0.00 & 0.00 & 0.20 \\
    Kimi K3 & 0.16 & 0.42 & 0.39 & 0.02 & 0.01 & 0.00 & 0.00 & 0.18 \\
    DeepSeek-V4.1-Flash & 0.10 & 0.51 & 0.32 & 0.01 & 0.03 & 0.01 & 0.01 & 0.15 \\
    \bottomrule
  \end{tabular}
\end{table}

\begin{table}[H]
  \caption{Per-encoding recall on the real-footage curve question, with columns in descending order of the number of real clips: D-Log (54), Log3G10 (53), F-Log2~C (31), S-Log3 (30), Apple Log~2 (AL\,2; 30), LogC3 (25), LogC4 (22), C-Log3 (21), C-Log2 (20), Film Gen5 (Gen5; 16), N-Log (12), Apple Log (AL; 11) and KineLOG3 (Kine; 5).
Bold marks the best value in each column within each prompt tier.}
  \label{tab:percurve}
  \centering
  \small
  \setlength{\tabcolsep}{1.5pt}
  \begin{tabular}{@{}lccccccccccccc@{}}
    \toprule
    Model & D-Log & Log3G10 & F-Log2\,C & S-Log3 & AL\,2 & LogC3 & LogC4 & C-Log3 & C-Log2 & Gen5 & N-Log & AL & Kine \\
    \midrule
    \multicolumn{14}{@{}l}{\emph{Frames only}} \\
    Claude Opus 5.5 & 0.34 & \textbf{0.11} & 0.00 & 0.47 & 0.00 & 0.10 & 0.00 & 0.00 & 0.00 & 0.00 & 0.00 & 0.18 & 0.00 \\
    Claude Sonnet 5.5 & 0.60 & 0.01 & 0.00 & 0.10 & 0.00 & \textbf{0.32} & 0.07 & 0.00 & 0.00 & 0.03 & 0.00 & \textbf{0.41} & 0.00 \\
    GPT-6.1 Sol & 0.58 & 0.05 & 0.00 & 0.33 & 0.00 & 0.02 & 0.02 & \textbf{0.07} & 0.00 & 0.00 & 0.00 & 0.05 & 0.00 \\
    GPT-6 Astra & 0.58 & 0.03 & 0.00 & 0.37 & 0.00 & 0.00 & 0.02 & 0.02 & 0.00 & 0.00 & 0.00 & 0.05 & 0.00 \\
    Gemini 3.1 Pro & 0.74 & 0.00 & 0.00 & 0.42 & 0.00 & 0.04 & 0.11 & 0.00 & \textbf{0.10} & 0.00 & \textbf{0.08} & 0.27 & 0.00 \\
    Gemini 3.8 Flash & \textbf{0.78} & 0.05 & 0.00 & 0.48 & 0.00 & 0.00 & \textbf{0.14} & 0.00 & 0.00 & 0.00 & 0.00 & 0.32 & \textbf{0.10} \\
    Qwen3.8-Max & 0.01 & 0.00 & 0.00 & 0.48 & 0.00 & 0.14 & 0.00 & 0.00 & 0.00 & 0.03 & 0.00 & 0.00 & 0.00 \\
    Qwen3.8-Flash & 0.39 & 0.00 & 0.00 & \textbf{0.75} & 0.00 & 0.06 & 0.02 & 0.00 & 0.00 & 0.03 & 0.00 & 0.00 & 0.00 \\
    Kimi K3.1 Pro & 0.73 & 0.03 & 0.00 & 0.60 & 0.00 & 0.08 & 0.02 & 0.00 & 0.00 & \textbf{0.09} & 0.00 & 0.14 & 0.00 \\
    Kimi K3 & 0.68 & 0.02 & 0.00 & 0.57 & 0.00 & 0.16 & 0.00 & 0.00 & 0.00 & 0.03 & 0.00 & 0.09 & 0.00 \\
    DeepSeek-V4.1-Flash & 0.45 & 0.00 & \textbf{0.03} & 0.52 & \textbf{0.02} & 0.18 & 0.02 & 0.00 & 0.00 & \textbf{0.09} & 0.00 & 0.00 & 0.00 \\
    \midrule
    \multicolumn{14}{@{}l}{\emph{Frames + spec + guide}} \\
    Claude Opus 5.5 & 0.56 & 0.04 & 0.00 & \textbf{0.62} & 0.00 & 0.02 & 0.02 & 0.00 & 0.00 & 0.00 & 0.00 & 0.27 & 0.00 \\
    Claude Sonnet 5.5 & 0.62 & 0.00 & 0.00 & 0.48 & 0.00 & \textbf{0.16} & 0.02 & 0.00 & 0.00 & 0.00 & 0.00 & 0.14 & 0.00 \\
    GPT-6.1 Sol & 0.69 & 0.34 & 0.00 & 0.25 & 0.00 & \textbf{0.16} & \textbf{0.20} & \textbf{0.40} & 0.00 & 0.00 & 0.00 & 0.32 & 0.00 \\
    GPT-6 Astra & 0.75 & \textbf{0.45} & 0.00 & 0.23 & 0.00 & 0.14 & \textbf{0.20} & 0.31 & 0.00 & 0.03 & 0.00 & 0.14 & 0.00 \\
    Gemini 3.1 Pro & 0.50 & 0.07 & 0.00 & 0.12 & 0.00 & 0.06 & 0.00 & 0.26 & 0.00 & 0.06 & 0.00 & 0.45 & 0.00 \\
    Gemini 3.8 Flash & 0.71 & 0.10 & 0.00 & 0.57 & 0.00 & 0.00 & 0.14 & 0.05 & 0.00 & 0.00 & 0.00 & 0.14 & 0.00 \\
    Qwen3.8-Max & 0.11 & 0.00 & 0.00 & 0.08 & 0.00 & 0.06 & 0.00 & 0.00 & 0.00 & 0.00 & 0.04 & \textbf{0.64} & 0.00 \\
    Qwen3.8-Flash & 0.63 & 0.01 & 0.00 & 0.38 & \textbf{0.03} & 0.02 & 0.00 & 0.00 & 0.00 & 0.03 & 0.00 & 0.41 & 0.00 \\
    Kimi K3.1 Pro & \textbf{0.81} & 0.01 & 0.00 & 0.45 & 0.00 & 0.06 & 0.05 & 0.02 & 0.00 & 0.12 & 0.00 & 0.32 & 0.00 \\
    Kimi K3 & 0.62 & 0.00 & 0.00 & 0.42 & 0.00 & 0.02 & 0.00 & 0.00 & 0.00 & 0.00 & \textbf{0.12} & 0.36 & 0.00 \\
    DeepSeek-V4.1-Flash & 0.67 & 0.00 & 0.00 & 0.35 & 0.02 & 0.14 & 0.00 & 0.00 & 0.00 & \textbf{0.22} & 0.00 & 0.41 & 0.00 \\
    \bottomrule
  \end{tabular}
\end{table}

\begin{table}[H]
  \caption{Per-brand recognition, confusion between similar encodings of the same brand, and the decoupled brand--curve subset, with frames only.
The first four columns report recognition accuracy on the brand question for the 330 real clips, grouped by the true brand: DJI (54 clips), Sony (30), Apple (41) and the other 7 brands together (205).
The middle three columns report, on the regular curve question, the share of answers that mistake the true curve for another encoding of the same brand: D-Log for D-Log~M (54 clips), Apple Log~2 for Apple Log (30), and LogC4 for LogC3 (22).
The last two columns report curve and brand accuracy on the decoupled subset of 35 RED Log3G10 clips shot on Nikon bodies.
The last row pools the 11 models.
Bold marks the highest value in each accuracy column and the highest share in each confusion column.}
  \label{tab:brand}
  \centering
  \small
  \setlength{\tabcolsep}{3pt}
  \begin{tabular}{@{}lccccccccc@{}}
    \toprule
    & \multicolumn{4}{c}{Brand recognition accuracy $\uparrow$} & \multicolumn{3}{c}{Confusion with a same-brand encoding} & \multicolumn{2}{c}{Log3G10 on Nikon} \\
    \cmidrule(lr){2-5}\cmidrule(lr){6-8}\cmidrule(l){9-10}
    Model & DJI & Sony & Apple & Other 7 & \shortstack{D-Log\\$\to$ D-Log M} & \shortstack{Apple Log 2\\$\to$ Apple Log} & \shortstack{LogC4\\$\to$ LogC3} & Curve $\uparrow$ & Brand $\uparrow$ \\
    \midrule
    Claude Opus 5.5 & 0.79 & \textbf{0.92} & 0.01 & 0.02 & 0.35 & 0.15 & 0.02 & \textbf{0.13} & 0.00 \\
    Claude Sonnet 5.5 & 0.77 & 0.67 & 0.00 & 0.04 & 0.07 & \textbf{0.43} & \textbf{0.32} & 0.00 & 0.00 \\
    GPT-6.1 Sol & 0.65 & 0.52 & 0.00 & 0.00 & 0.06 & 0.10 & 0.00 & 0.06 & 0.00 \\
    GPT-6 Astra & 0.62 & 0.38 & 0.00 & 0.00 & 0.06 & 0.12 & 0.00 & 0.04 & 0.00 \\
    Gemini 3.1 Pro & \textbf{0.81} & 0.43 & 0.04 & \textbf{0.07} & 0.00 & 0.28 & 0.05 & 0.00 & \textbf{0.07} \\
    Gemini 3.8 Flash & 0.69 & 0.50 & 0.00 & 0.04 & 0.02 & 0.23 & 0.00 & 0.04 & 0.00 \\
    Qwen3.8-Max & \textbf{0.81} & 0.47 & 0.02 & 0.03 & \textbf{0.59} & 0.02 & 0.07 & 0.00 & 0.00 \\
    Qwen3.8-Flash & 0.61 & 0.58 & 0.00 & 0.05 & 0.30 & 0.00 & 0.02 & 0.00 & 0.00 \\
    Kimi K3.1 Pro & 0.69 & 0.58 & \textbf{0.05} & 0.03 & 0.05 & 0.18 & 0.09 & 0.04 & 0.00 \\
    Kimi K3 & 0.68 & 0.70 & \textbf{0.05} & 0.05 & 0.01 & 0.28 & 0.23 & 0.01 & 0.00 \\
    DeepSeek-V4.1-Flash & 0.75 & 0.63 & 0.02 & 0.02 & 0.13 & 0.08 & 0.14 & 0.00 & 0.00 \\
    \midrule
    11 models pooled & 0.71 & 0.58 & 0.02 & 0.03 & 0.15 & 0.17 & 0.08 & 0.03 & 0.01 \\
    \bottomrule
  \end{tabular}
\end{table}

\section{Ablations on input format and number of frames}
\label{app:ablation}

\noindent\textbf{Bit depth and file format.}
On the 550 Log images of the synthetic natural-scene set, the formula baseline reaches accuracies of 0.71, 0.71 and 0.68 with the 16-bit originals, 8-bit lossless images and 8-bit JPEG images as input; the three formats differ by no more than 3 percentage points.

\noindent\textbf{Joint multi-frame input versus independent single-frame input.}
When the three frames of each real clip are separated and asked independently (\Cref{tab:frames}), the single-frame accuracy at the 1/4, 1/2 and 3/4 time points differs by at most 0.02.
Compared with the joint three-frame input, GPT-6.1 Sol (0.14 $\to$ 0.08--0.09) and Kimi K3.1 Pro (0.20 $\to$ 0.14--0.16) drop, while Claude Opus 5.5 (0.13 $\to$ 0.14--0.15) and Qwen3.8-Max (0.06 $\to$ 0.04--0.06) stay about level.
The three single-frame questions of one clip receive the same answer in only 0.17--0.62 of the cases, so a single-frame prediction depends considerably on the sampled position, and the joint three-frame input provides a more stable clip-level representation.

\begin{table}[H]
  \caption{Frame-sampling ablation with frames only, on the 330 real Log clips.
Curve-identification accuracy with the joint three-frame input and with independent single frames at 1/4, 1/2 and 3/4 of the duration, and the share of clips whose three single-frame questions receive the same answer.
Bold marks the best value in each column.}
  \label{tab:frames}
  \centering
  \setlength{\tabcolsep}{3pt}
  \begin{tabular}{@{}lccccc@{}}
    \toprule
    & Joint & \multicolumn{3}{c}{Single frame} & Same \\
    \cmidrule(lr){3-5}
    Model & 3 frames & 1/4 & 1/2 & 3/4 & answer \\
    \midrule
    Claude Opus 5.5 & 0.13 & \textbf{0.15} & 0.14 & 0.14 & 0.48 \\
    GPT-6.1 Sol & 0.14 & 0.09 & 0.08 & 0.08 & \textbf{0.62} \\
    Qwen3.8-Max & 0.06 & 0.04 & 0.04 & 0.06 & 0.17 \\
    Kimi K3.1 Pro & \textbf{0.20} & 0.14 & \textbf{0.15} & \textbf{0.16} & 0.31 \\
    \bottomrule
  \end{tabular}
\end{table}

\section{Prompt templates and probe options}
\label{app:prompts}

All items share one English prompt template; only two placeholders, the reference material and the options, are replaced according to the prompt tier and the question type, and they are marked with square brackets in the template below.
\Cref{tab:tiers} lists the inputs and the diagnostic purposes of the four prompt tiers.

\begin{table}[H]
  \caption{Inputs and diagnostic purposes of the four prompt tiers. We evaluate real items on all tiers except \emph{frames + spec}, and synthetic items on the three visual tiers.}
  \label{tab:tiers}
  \centering
  \begin{tabular}{@{}lp{0.31\textwidth}p{0.41\textwidth}@{}}
    \toprule
    Tier & What the model receives & What it evaluates \\
    \midrule
    Text only & Question and options, no images & Whether the wording of the question and the options introduces a bias \\
    Frames only & Lossless frames, question and options & Identification from the model's internal knowledge alone \\
    Frames + spec & Adds the specification digest with formulas and key code values only & Whether the model can measure pixels by itself once it knows the exact parameters \\
    Frames + spec + guide & Further adds the six-step measurement procedure & Identification once the measurement method is supplied as well \\
    \bottomrule
  \end{tabular}
\end{table}

\noindent\textbf{System prompt.}
\begin{lstlisting}
You are a senior DIT and colorist with deep, hands-on knowledge of camera log encodings, their transfer functions and native colour gamuts. When images are attached, each one is an 8-bit screenshot of a video frame exactly as the file stores it: we applied no LUT, colour-space transform, grade or display transform of our own, so pixel values are the file's code values scaled down to 8 bits (multiply by about 4 to get 10-bit code values on a 0-1023 scale). The file itself may be a camera-original log recording, or it may already be display-referred (e.g. Rec.709 / a baked-in look); deciding which is part of the task. Answer strictly with a single JSON object and nothing else.
\end{lstlisting}

\noindent\textbf{User prompt.}
The template below is for the curve question and the probe; synthetic items replace the first sentence with a description of a single image.
The brand question replaces the question with ``Which camera brand most likely recorded this footage? Also name up to three specific camera models, most likely first, in "camera\_candidates".'', replaces the options with the 11 brands and adds the output field \texttt{camera\_candidates}.
\begin{lstlisting}
3 frames from the same clip are attached, in time order, taken at 25%, 50%, 75% of its duration. They share one encoding; use all of them.

Reference material: below is a digest of the camera manufacturers' white papers (encoding formulas and key code values). Use its numeric cues to check your reading of the frame.
<reference>
[REFERENCE MATERIAL]
</reference>

Question: Which camera log encoding (transfer function) and native colour gamut was this footage most likely recorded in?

Work in this order:
1. Before reading the options, write your own free-text best guess of the encoding and gamut in "open_answer".
2. List the cues you relied on in "evidence". Be quantitative wherever you can: estimated black level / darkest code value, where a neutral mid-grey or skin tone sits, highlight clip or plateau code value, hue bias of strongly saturated colours. Give code values on the 10-bit scale (0-1023) or normalised 0-1.
3. Pick your two most likely options ("top2", most likely first) and your single final answer ("choice", normally the first entry of "top2"). Use option letters only.

Options:
[OPTIONS]

Respond with a single JSON object and nothing else:
{"open_answer": "<free-text best guess, made before reading the options>",
 "evidence": ["<observable cue 1>", "<cue 2>", "..."],
 "top2": ["<letter>", "<letter>"],
 "choice": "<single letter>"}
\end{lstlisting}

\noindent\textbf{Reference material.}
The frames-only and text-only tiers contain no \texttt{Reference material} block.
The frames + spec tier fills in sections 1--5 of the specification digest, which cover the key code values, the formulas, the gamut primaries, the clipping values by sensitivity and the sources.
The frames + spec + guide tier further attaches sections 6--8, reproduced verbatim below.
\begin{lstlisting}
## 6. What actually separates the encodings

- **Black level families**: ~95 = ARRI LogC3 95, ARRI LogC4 95, Sony S-Log3 95, DJI D-Log 95, RED Log3G10 94, BMD Film Gen 5 95, Fujifilm F-Log2 95, Kinefinity KineLOG3 95, Canon Log 2 95, Fujifilm F-Log2 C 95; ~128 = Canon Log 3 128, Nikon N-Log 127; ~154 = Apple Log 154, Apple Log 2 154. Only true black (lens cap, deep unlit shadow) shows it; a scene floor above the black level proves nothing, a floor clearly *below* it rules the curve out.
- **18% grey**: ARRI LogC4 285 < RED Log3G10 341 < Canon Log 3 351 < Nikon N-Log 372 < BMD Film Gen 5 392 < ARRI LogC3 400 < Fujifilm F-Log2 400 < Fujifilm F-Log2 C 400 < Kinefinity KineLOG3 401 < Canon Log 2 407 < DJI D-Log 408 < Sony S-Log3 420 < Apple Log 500 < Apple Log 2 500. Needs a grey card or a confidently known mid-tone; an exposure error of +-1 stop moves grey by 57-85 CV, more than most gaps.
- **Contrast (CV per stop at grey)**: ARRI LogC4 57, BMD Film Gen 5 60, RED Log3G10 63, Canon Log 2 70, Fujifilm F-Log2 71, Fujifilm F-Log2 C 71, ARRI LogC3 72, DJI D-Log 74, Sony S-Log3 74, Kinefinity KineLOG3 76, Nikon N-Log 83, Apple Log 83, Apple Log 2 83, Canon Log 3 85. D-Log M (community fit) is ~112, i.e. not a true log.
- **Clip behaviour**: LogC3 and D-Log clip levels move with EI (section 4). Log3G10 reaches 1.0 only at 184 (10 stops over grey), BMD Gen 5 at 223, LogC4 at 470, so their highlights usually clip well below 1023 at a camera-specific level.
- **Gamut (blue primary y)**: Kinefinity WG -0.224, AWG3 -0.102, Cinema Gamut -0.100, S-Gamut3.Cine -0.087, RWG -0.085, BMD WG -0.082, D-Gamut -0.080, S-Gamut3 -0.050, AWG4 -0.031, Rec.2020 +0.046. RWG (green y 1.494), Kinefinity WG (1.148) and Cinema Gamut (1.14) contain large imaginary regions. The same saturated object therefore has different RGB code ratios in each encoding.
- **EI dependence**: only LogC3 changes the curve itself (section 4); D-Log changes only the clip level; all others are single curves.

## 7. How a DIT uses these numbers

1. Read code values with a 10-bit full-range scope (0-1023). If blacks sit near 64 and nothing exceeds 940, the file may have been squeezed to video levels: full-range code y then appears as 64 + 0.856 * y (black 95 -> 145, grey 400 -> 406); BMD camera ProRes is recorded this way.
2. Find the darkest flat area (lens cap, deep shadow under an object). A floor at 93-96 points to the 95-family, 126-130 to Canon Log 3 / N-Log, ~154 to Apple Log. Noise makes single pixels dip below black; use averages over flat patches.
3. Find a neutral reference (grey card, white card, overcast sky is NOT neutral-reliable). Compare its code with the grey / white columns, remembering the exposure is unknown: use differences between two known patches (e.g. grey vs. white = 2.32 stops) to estimate CV/stop, which is exposure-independent.
4. Look for a hard highlight plateau (many pixels at one identical value, often in all channels). Compare its code with the documented clip levels (section 4). A plateau above a curve's maximum possible code rules that curve out.
5. Decode saturated, well-exposed patches with each candidate curve + gamut to XYZ. Physically real surface colours stay inside the spectral locus; many negative or out-of-locus results mean the assumed gamut is too wide or the curve too steep. Too-small saturation everywhere hints the assumed gamut is too narrow.
6. Combine: black family + grey/CV-per-stop + clip level + gamut plausibility. Each cue alone is weak; when they disagree, report the top-2 candidates instead of a single answer.

Worked example (studio chart): averaged floor 128 on a lens-cap black patch, a grey card at ~372, and no plateau anywhere -> the 128-family (Canon Log 3 / N-Log); grey 372 matches N-Log (Canon Log 3 would put grey at 351; check with the white card: N-Log 603, Canon Log 3 577). Same chart with floor 95 and grey 400 but no clip plateau: the 95-family stays open (LogC3 / F-Log2 first by grey, then S-Log3 420, D-Log 408, BMD 392, Log3G10 341, LogC4 285 only if the image is clearly under-exposed).

## 8. Caveats

- D-Log M is visibly higher-contrast than D-Log.
- Container colour tags (e.g. Rec.709 nclc) and re-exports say nothing about the original encoding.
\end{lstlisting}

\noindent\textbf{Options.}
The probe uses the same frames and prompt as the regular curve question and only removes the correct encoding from the options.
For a question whose correct answer is RED Log3G10, the table below gives the two option lists under the first option order.

\begin{center}
  \begin{tabular}{@{}l>{\raggedright\arraybackslash}p{2.6cm}l@{}}
    \toprule
    Curve question & Probe & Option \\
    \midrule
    A & A & ARRI LogC3 / ALEXA Wide Gamut (AWG3) \\
    B & B & ARRI LogC4 / ALEXA Wide Gamut 4 (AWG4) \\
    C & C & Sony S-Log3 / S-Gamut3.Cine \\
    D & D & DJI D-Log / D-Gamut \\
    E & E & DJI D-Log M \\
    \textbf{F} (answer) & removed; later letters move up by one & \textbf{RED Log3G10 / REDWideGamutRGB} \\
    G & F & Canon C-Log2 / Cinema Gamut \\
    H & G & Canon C-Log3 / Cinema Gamut \\
    I & H & Blackmagic Film Gen 5 / Blackmagic Wide Gamut Gen 5 \\
    J & I & Apple Log / Rec.2020 \\
    K & J & Apple Log 2 / Apple Wide Gamut \\
    L & K & Fujifilm F-Log2 / F-Gamut (Rec.2020 primaries) \\
    M & L & Fujifilm F-Log2 C / F-Gamut C \\
    N & M & Nikon N-Log / Rec.2020 \\
    O & N & Kinefinity KineLOG3 / Kinefinity Wide Gamut \\
    P & O & Not a Log encoding (already display-referred, e.g. Rec.709 / sRGB) \\
    Q & \textbf{P} (answer) & Cannot be determined from this image \\
    \bottomrule
  \end{tabular}
\end{center}

\end{document}